\documentclass{article}

\usepackage[preprint]{neurips_2026}

\usepackage{microtype}
\usepackage{graphicx}
\usepackage{subcaption}
\usepackage{booktabs} 
\usepackage{hyperref}
\usepackage{mathtools}
\usepackage{amsthm}

\usepackage{listings}   
\usepackage{xcolor}
\usepackage{xr}
\usepackage{amsmath}
\usepackage{microtype}
\usepackage{graphicx}
\usepackage{url}
\usepackage{physics}
\usepackage{booktabs} 
\usepackage{multirow}
\usepackage{colortbl}
\usepackage{rotating}
\usepackage{enumitem}

\usepackage{lipsum}
\usepackage{subcaption}
\usepackage{xspace}
\usepackage{algpseudocode}
\usepackage{amssymb}
\usepackage{pifont}

\usepackage{makecell}
\usepackage{tcolorbox}

\usepackage{caption} 

\lstdefinestyle{cvprprompt}{
  basicstyle=\ttfamily\small,   
  breaklines=true,
  breakatwhitespace=true,
  frame=single,
  columns=fullflexible,
  numbers=none,                 
  xleftmargin=0em,
  xrightmargin=0em,
}

\usepackage[textsize=tiny]{todonotes}

\title{GS-Agent: Creating 4D Physical Worlds\\With Generative Simulation}

\author{
  Hongxin Zhang$^{1,2}$, Chunru Lin$^{1}$, Junyan Li$^{1}$, Zhou Xian$^{2}$, Tsun-Hsuan Wang$^{2}$, Chuang Gan$^{1}$ \\
  $^1$ University of Massachusetts Amherst $^2$ Genesis AI \\
  \texttt{\{hongxinzhang,chunrulin,junyanli,chuangg\}@umass.edu} \\
  \texttt{\{zhouxian,johnson\}@genesis-ai.company}
}
    
\begin{document}

\maketitle

\vspace{-8mm}
\begin{minipage}{\textwidth}
    \centering
    \includegraphics[trim=000mm 000mm 000mm 000mm, clip=False, width=\linewidth]{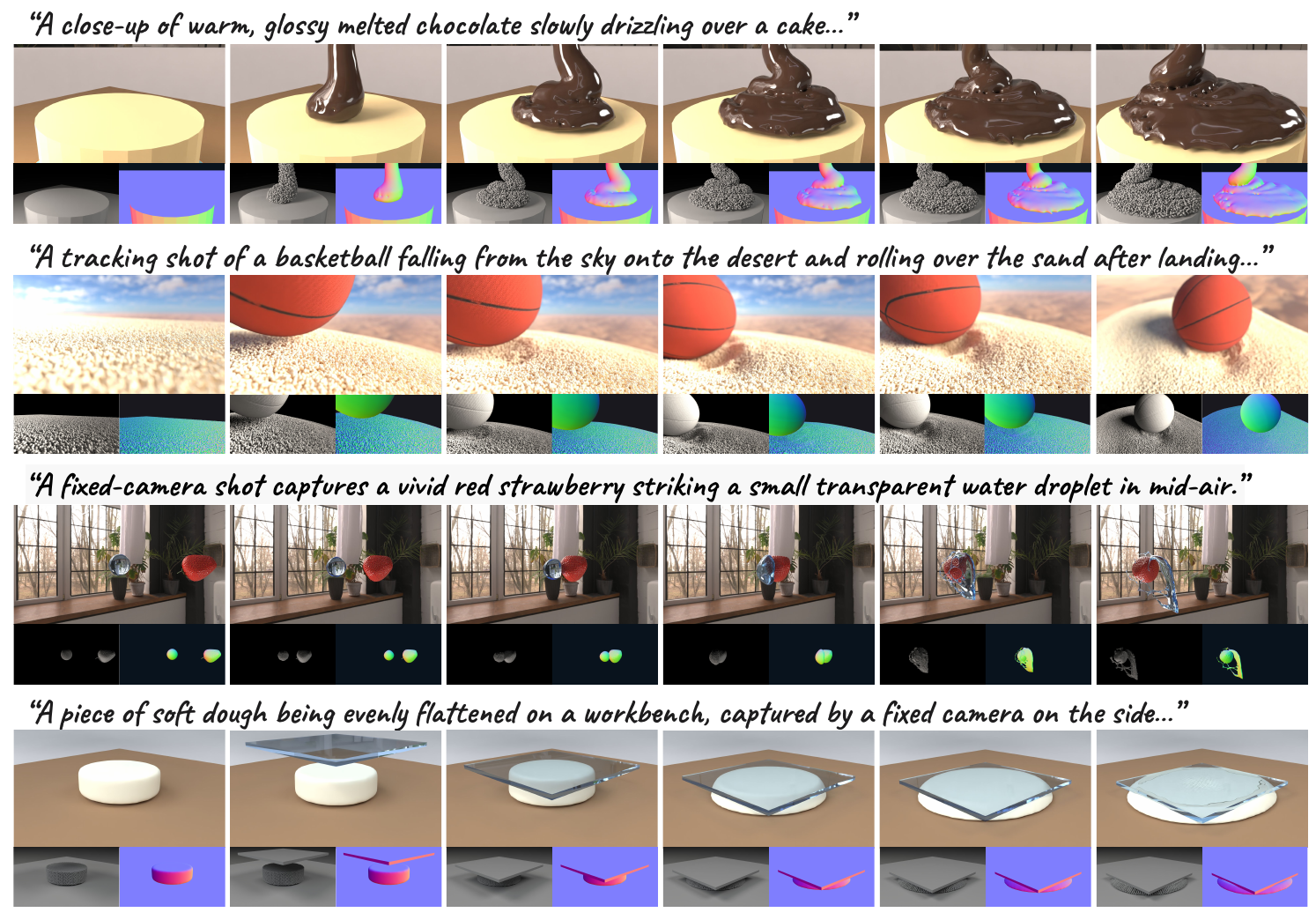}
\end{minipage}
\captionsetup{type=figure}
\captionof{figure}{\textbf{GS-Agent} creates 4D worlds from natural language through generative simulations, producing \textbf{physically plausible interactions} among liquids, deformable objects, and rigid bodies, together with \textbf{cinematic camera and lighting controls}. GS-Agent generates more than pixels; data modalities including fine-grained dynamics, depth and surface normals enable richer applications.}
\label{fig:teaser}
\vspace{1em}
    
\begin{abstract}
Creating dynamic and physically realistic 4D worlds from natural language descriptions is both fascinating and challenging. Traditional computer graphics methods rely on manual creation, requiring extensive human effort to fine-tune materials, motions, and visual fidelity. Recent advances in generative foundation models have sparked interest in learning to generate such 4D worlds from large-scale data; however, existing methods still struggle to ensure physical plausibility and controllability.
In this work, we take a different path by leveraging foundation models to construct an agentic system that emulates how humans traditionally create 4D worlds, yet automates the entire process. We present GS-Agent, an end-to-end multi-agent framework that integrates physics engines in the loop to generate realistic, dynamic, and controllable 4D physical worlds from natural language. Inspired by how humans build 4D worlds, GS-Agent decomposes the task into entity management, covering 3D asset curation, material tuning, placement, and motion control, and rendering configuration, including camera and lighting manipulation. Multiple agents with distinct expertise interact with the physics engine via code, seek multimodal feedback, and collaborate to iteratively construct 4D worlds that align with the given descriptions.
Experimental results show that GS-Agent effectively converts natural language into diverse and physically plausible 4D worlds exhibiting rich interactions among liquids, deformable objects, and rigid bodies, while achieving cinematic camera and lighting control. 
We envision GS-Agent as a foundation for a new paradigm in 4D world generation, empowering creative content creation and physical AI. See our project page\footnote{\url{https://umass-embodied-agi.github.io/gs-agent/}} for videos.
\end{abstract}

\section{Introduction}
\label{sec:intro}

Creating dynamic, physically plausible 4D worlds has far-reaching applications across domains such as embodied AI, autonomous driving, gaming, and film production. The ability to construct diverse and realistic physical environments enables embodied agents to be trained and evaluated in safer, richer settings, helping to narrow the sim-to-real gap during deployment~\citep{zhao2020sim}. Moreover, allowing users to build and control 4D worlds through natural language dramatically lowers the barrier to creative content creation, opening the door to more expressive storytelling, richer entertainment experiences, and entirely new creative formats. However, traditional approaches require tremendous human effort, from curating 3D assets and tuning material parameters to constructing scenes, orchestrating object motions, setting up lighting, and designing camera trajectories.

Recent advances in foundation models have significantly improved language understanding~\citep{openai2023gpt4}, image synthesis~\citep{liu2023visual, ho2020denoising}, video generation~\citep{blattmann2023stable, yangcogvideox, sora2, wan2.2}, and 3D/4D scene creation~\citep{yu2024wonderjourney, bah20244dfy}. Yet these models still struggle with physical plausibility~\citep{meng2024towards, kang2024far} and controllability~\citep{he2024cameractrl}. These limitations suggest that learning 4D worlds solely from large-scale data remains insufficient to produce consistent, physically grounded scenes. A complementary direction, inspired by how humans create worlds, is to leverage the reasoning capabilities of large language models~\citep{guo2025deepseek} to build agentic systems that plan, critique, and iteratively refine world generation. Prior work has shown the promise of agentic systems: \cite{yang2024swe} develops a software-engineering agent capable of solving GitHub issues autonomously, and \cite{hu2024scenecraft} constructs an agentic pipeline for crafting static 3D scenes through Blender.

In this work, we introduce \textbf{GS-Agent}, an end-to-end multi-agent framework that integrates physics engines in the loop to generate realistic, dynamic, and controllable 4D physical worlds directly from natural language descriptions. Inspired by human workflows, GS-Agent decomposes the creation process into two coordinated components: \textit{entity management}, which includes 3D asset retrieval, material tuning, object placement, and motion control, and \textit{rendering configuration}, which covers camera manipulation and lighting design. Multiple specialized agents interact with the physics engine through code, obtain multimodal feedback, and collaboratively refine the simulation to produce worlds that faithfully follow the given instructions.

We evaluate GS-Agent on scenes involving complex interactions among liquids, deformable objects, and rigid bodies, as well as diverse camera and lighting controls. Compared to state-of-the-art text-to-video models (Sora2~\citep{sora2} and Wan2.2~\citep{wan2.2}) and agentic baselines (SWE-Agent~\citep{yang2024swe} and our enhanced variant with visual feedback), GS-Agent generates 4D worlds that demonstrate higher physical plausibility, stronger instruction fidelity, and finer controllability. Moreover, its agentic design allows it to identify and address unexpected failures in a fully autonomous manner. We envision GS-Agent as a stepping stone toward a new paradigm for 4D world generation, enabling interactive content creation and supporting large-scale, physically grounded data generation.
In summary, our contributions are threefold:

\begin{itemize}[leftmargin=*, itemsep=0mm]
    \item We tackle the challenging problem of 4D physical world creation from natural language by leveraging powerful physics engines with an automated agentic framework that emulates human 4D world-building workflows.
    \item We introduce GS-Agent, an end-to-end multi-agent framework that interacts with the physics engine via code, actively seeks multi-modal feedback, and collaboratively constructs 4D worlds that align with the natural language.
    \item Experimental results demonstrate that GS-Agent produces more physically plausible, controllable, and instruction-aligned 4D worlds than existing text-to-video and agentic baselines, highlighting the potential of combining agentic frameworks with powerful physics engines.
\end{itemize}
\section{Related Work}

\subsection{3D and 4D Generation}
\vspace{-1mm}
3D scene creation from natural languages has been a long-standing challenge~\citep{chang2014learning}. While early works rely on rule-based language-spatial mapping~\citep{chang2015text, ma2018language}, modern approaches learn to generate 3D contents from massive data~\citep{yu2024wonderjourney, yu2025wonderworld, chung2023luciddreamer, wang2024architect, lu2025ll3m}, mostly relying on generative diffusion models~\citep{wang2024diffusion}, and Gaussian Splatting~\citep{Xie_2024_CVPR}. To incorporate physically plausible motions into 3D scenes, many works leverage physics solvers~\citep{chen2025physgen3d, bah20244dfy, xu2024comp4d, zhang2024physdreamer, gillmanforce, wangphysctrl}. Notably, \cite{lv2024gpt4motion} leverages LLM to write Blender scripts for motion control, then generates video conditioned on physics-plausible frames. \cite{li2025wonderplay} generates action-conditioned dynamic 3D scenes from a single image with a hybrid generative simulator. Differently, we propose to build end-to-end agentic systems to automate the process of creating 4D worlds, which do not suffer from unreliable generalization of the visual generative models.

\vspace{-2mm}
\subsection{Text-to-Video Generation}
\vspace{-1mm}

Modern video generation models are mainly built on diffusion models~\citep{sohl2015deep, ho2020denoising, blattmann2023stable, ho2022imagen, chen2024videocrafter2} or auto-regressive models~\citep{yu2023magvit, kondratyuk2023videopoet}. The introduction of Diffusion Transformer (DiT)~\citep{peebles2023scalable, sora2, yangcogvideox} further demonstrates the potential of learning to generate video by scaling data, and some work incorporates physics priors to improve physics realism of the video~\citep{yuan2025newtongen, lv2024gpt4motion, liu2024physgen}. However, these models still struggle with physical plausibility~\citep{kang2024far}, controllability~\citep{he2024cameractrl, Zhang_2025_CVPR}, and consistency~\citep{bruce2024genie}. Our method crafts the world in a physics engine that provides plausible physical interactions.

\vspace{-2mm}
\subsection{Foundation Model Agents}
\vspace{-1mm}

With the rapid advancement of Foundation Models~\citep{openai2023gpt4, guo2025deepseek}, a surge of powerful agents has emerged~\citep{sumers2023cognitive, xi2023rise,wang2023survey}. These range from agents operating solely in the text domain~\citep{gur2023real, shinn2024reflexion} to multi-modal agents capable of interacting with GUIs~\citep{hong2024cogagent, agashe2025agent}, functioning as robots in the physical world~\citep{ahn2022can, huang2023voxposer, du2023video}, and general-purpose agents~\citep{qiu2025alita, hu2025owl}. Multi-agent frameworks have shown promising performance gain over various tasks~\citep{li2023camel, wu2024autogen, hu2025owl}, including embodied AI~\citep{zhang2024building}, reasoning~\citep{du2023improving}, and code generatoin~\citep{liu2024a}. Notably, \cite{yang2024swe} builds a software engineering agent that can solve real github issues automatically, \cite{hu2024scenecraft, sun20253d, yin2026vision,liuir3d} build agents that can leverage Blender, a 3D software, to craft static 3D scenes. Different from them, we introduce the first agent that can build 4D worlds from natural languages with a unified physics engine in the loop.

\section{GS-Agent: A Multi-Agent Framework with Generative Simulation in the Loop}

\begin{figure*}[tbp]
    \centering
    \includegraphics[width=0.9\linewidth]{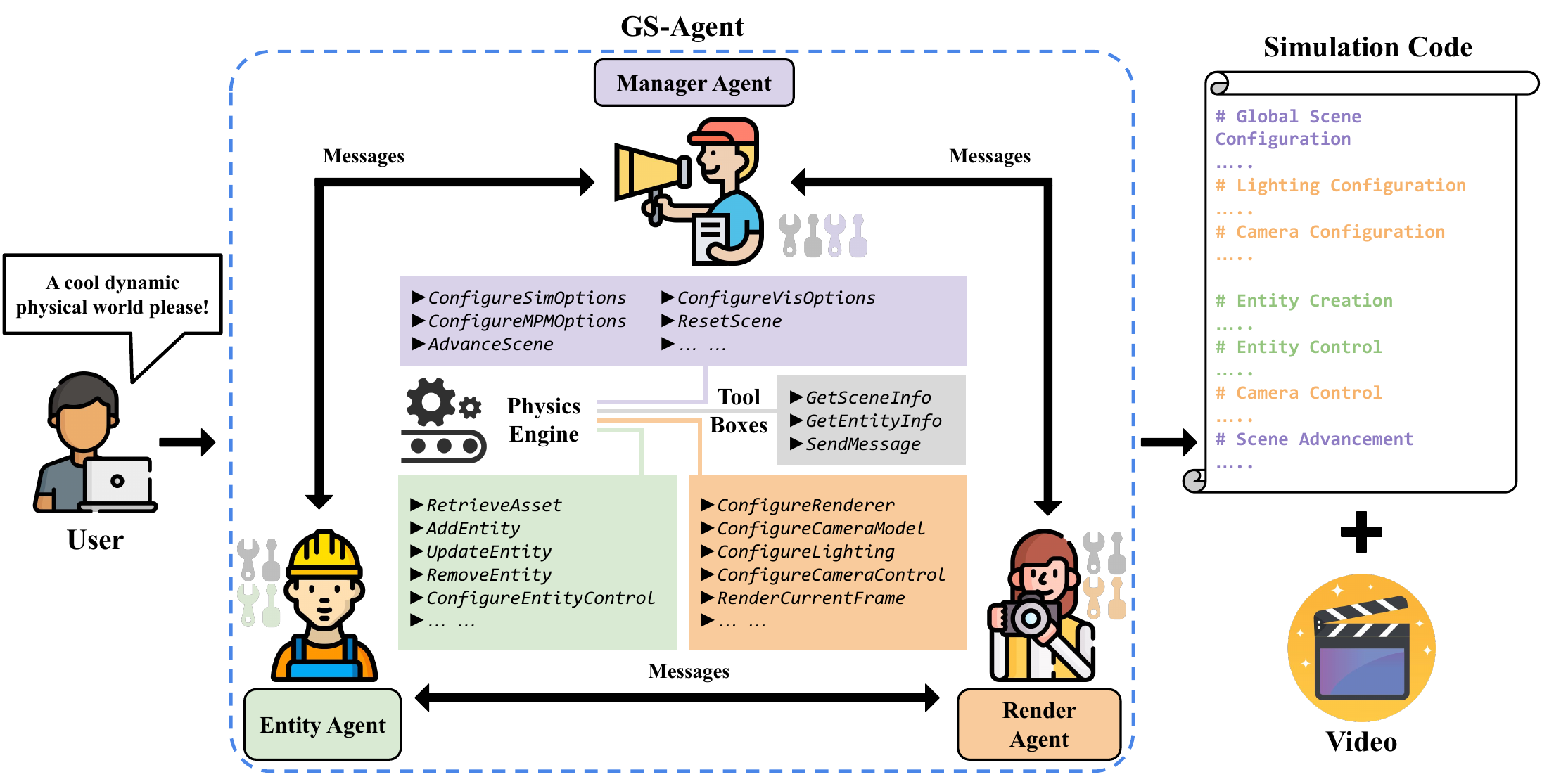}
    \caption{\textbf{Overview of GS-Agent.} The \textit{Manager Agent} interprets the natural language description, decomposes it into actionable sub-tasks, configures global simulation options, controls scene advancement, and delegates tasks to the appropriate agents. The \textit{Entity Agent} curates assets for morphology creation, places entities according to spatial constraints, tunes material parameters for physical plausibility, and controls entity motions. The \textit{Render Agent} manipulates lighting and camera settings, composes camera trajectories, and configures video recording. Agents communicate via a \textit{SendMessage} tool and interact with the physics engine through structured tool interfaces. The final 4D world is represented as executable simulation code and a rendered video. Full tool descriptions are provided in Appendix~\ref{app:method}.}
    \vspace{-6mm}
    \label{fig:framework}
\end{figure*}

Creating a 4D world is a complex, multi-stage process that typically requires coordinated effort across different areas of expertise: entity modeling, material design, spatial arrangement, motion specification, lighting setup, and camera planning. In human production pipelines, such as visual effects or game development, a director translates high-level requirements into actionable plans, delegates tasks to specialists, evaluates intermediate outputs, and coordinates iterative refinement through natural language communication. 
Inspired by this workflow, we introduce \textbf{GS-Agent}, an end-to-end multi-agent framework in which three specialized agents, the \textit{Manager Agent}, \textit{Entity Agent}, and \textit{Render Agent}, collaboratively construct a physically plausible 4D world upon a physics engine, as shown in Figure~\ref{fig:framework}. We first cover basic concepts of a \textit{physics engine} in sec~\ref{sec:engine}, then the three specialized agents: the Manager Agent handles user communication, step-by-step planning, task delegation, verification, and global simulation configuration in sec~\ref{sec:manager}; the Entity Agent specializes in entity creation, morphology, material assignment, and motion control in sec~\ref{sec:entity}; the Render Agent governs camera and lighting manipulation and provides visual feedback to the other agents in sec~\ref{sec:render}. Agents communicate with each other via a \textit{SendMessage} tool and interact with the physics engine through structured tool interfaces.
Underlying the system, all agents generate and refine executable code that interfaces with the physics engine, request multimodal feedback including physics boundary checks, program runtime information, images, and videos, communicate with each other, and iteratively improve the world until the natural language goals are satisfied. A complete list of prompts and tool interfaces is provided in Appendix~\ref{app:method}.

\subsection{Physics Engine}
\label{sec:engine}
A physics engine provides a unified computational framework for constructing and simulating dynamic worlds. Its core components include:
\begin{itemize}[leftmargin=*, itemsep=2mm]
    \vspace{-1mm}
    \item \textbf{Entity}. The entity component defines the objects in the scene. Each entity is defined as $E=(\textit{Morph}, \textit{Material}, \textit{Surface})$, where \textit{Morph} specifies geometry (primitive shape or mesh), \textit{Material} determines physical behavior (rigid, deformable, liquid), and \textit{Surface} controls visual attributes. It also includes control parameters such as initial velocities, applied forces, kinematic trajectories, and constraints, providing the solvers with fully specified object states and motion inputs.
    \vspace{-1mm}
    \item \textbf{Solver}. The solver component governs how the engine computes physical dynamics. Solvers are selected and configured based on entity materials, e.g., rigid-body solvers for solid objects, continuum methods such as Material Point Method (MPM)~\citep{jiang2016material} for deformable objects, and particle-based solvers like Smoothed Particle Hydrodynamics (SPH)~\citep{muller2003particle} for fluids. Solver parameters such as time step and resolution govern stability and accuracy, ensuring that entity motion follows consistent physical laws.
    \vspace{-1mm}
    \item \textbf{Renderer}. The renderer component displays the results of the physics simulation on the screen, and it is typically a system separate from the physics engine itself. Renderer parameters include the Camera (its position, angle, and field of view), Lighting (the type, position, color, and intensity of light sources) and Shadows (their quality and draw distance).
\end{itemize}

\subsection{Manager Agent}
\label{sec:manager}
\noindent\textbf{Task decomposition and delegation.}
Given a natural language description, the Manager Agent interprets the user intent, reasons about the required scene components, and formulates an actionable multi-step plan. It assigns one sub-task at a time to the Entity Agent or Render Agent based on their capabilities—for example, asset creation with appropriate geometry and material properties, spatial placement under geometric constraints, or camera trajectory refinement for a desired cinematic effect. After each sub-task is completed, the Manager Agent critically verifies the result and triggers iterative refinement until the specification is met.

\noindent\textbf{Scene Configuration.}
The Manager Agent configures the simulation and visualization options both before and during creation. It sets global simulation parameters (e.g., time step, substeps, gravity), solver-specific settings (e.g., MPM grid resolution and domain bounds, SPH particle size), and visualization options (shadows, background color, particle rendering mode, segmentation levels). These configurations ensure stable physics, consistent domain scales, and clear visual diagnostics for downstream reasoning.

\noindent\textbf{Timeline Control.} To assess the evolving dynamics of the constructed world, the Manager Agent regulates the temporal progression of the simulation. It can advance or reset the scene to probe intermediate states, evaluate the outcomes of modifications, or perform multi-stage rollouts. After verifying that all components satisfy the language specification, the agent initiates the final execution and triggers video capture, producing the completed 4D world.

\subsection{Entity Agent}
\label{sec:entity}

\noindent\textbf{3D Model Asset Retrieval.}
To construct realistic entities, the Entity Agent first tries to retrieve assets from a curated library (BlenderKit\footnote{\url{https://www.blenderkit.com/}} and PolyHaven\footnote{\url{https://polyhaven.com/}}) using a semantic index built from text and image embeddings. Query embeddings are produced using a text encoder and CLIP~\citep{radford2021learning}, concatenated and normalized before searching against precomputed metadata and multi-view image embeddings. The retriever returns top candidates with information helpful for downstream creation: local file path, short description, preview images, match score, and axis-aligned bounds.

\noindent\textbf{3D Model Asset Generation.} If the agent fails to retrieve appropriate 3D model assets that meet the requirement, it will try to generate the 3D model asset by calling Meshy~\citep{meshy2024}, a text-to-3D model, and calculate the same metadata of the assets and save it to the curated assets library. Based on the generation results, the agent may adjust the text query accordingly. After a maximum number of failed attempts, the agent will try to fall back to using primitives to construct the required morphologies.

\noindent\textbf{Entity Placement.}
Based on 3D model assets metadata, the Entity Agent computes plausible scale, orientation, and position. It aligns assets using canonical orientation, rescales meshes from AABB bounds, and places them to satisfy world-space constraints such as resting surfaces, non-interpenetration, and adequate spacing. The Entity Agent queries the scene state and ask Render Agent for visual feedback; if misalignment or collisions are detected, it corrects and iterates.

\noindent\textbf{Physically Plausible Material Tuning.}
Tuning physical material parameters is notoriously difficult and often impractical for non-experts~\citep{liu2012physicalMaterialEditing}. Misconfigured parameters can result in implausible or unstable simulations. The Entity Agent initializes material choices using common-sense priors, then iteratively refines parameters based on real physics feedback from the physics engine. Take continuum materials as an example, Entity Agent tunes principled parameters such as Young’s modulus, Poisson’s ratio, and yield stress (for plasticity) according to the actual degree of deformation exhibited in the simulation. It also aligns per-entity material parameters with solver constraints (e.g., MPM grid size, SPH particle radius) to maintain stable resolution scales.

\noindent\textbf{Entity Motion.}
The Entity Agent controls motion by editing per-step control functions executed during simulation. Supported controls include PD controllers, direct position/velocity/force commands for rigid bodies, and programmable emitters for particle-based materials.  This interface allows scripting kinematic trajectories and transient particle events within the same simulation timeline.

\subsection{Render Agent}
\label{sec:render}

\noindent\textbf{Camera Positioning.}
The Render Agent determines camera placement and intrinsics for both intermediate inspection and final rendering. It interprets verbal directives, such as requests for specific viewpoints or framings, by grounding them in the world coordinate system. When necessary, the agent queries the spatial state of entities (e.g., positions and extents) to position the camera in a manner consistent with the requested visual perspective.

\noindent\textbf{Camera Motion and Video Capturing.}
To generate temporally coherent visualizations, the Render Agent represents camera motion as explicit control code that is executed over the course of the simulation. This code specifies camera trajectories (e.g., orbits, dolly movements, or entity-tracking motions) and the corresponding frame-capture schedule. Because trajectories are coded rather than baked into pixels, they can be further refined, for example, through smoothing or reparameterization, without changing the underlying physical simulation. Camera updates are synchronized with simulation time steps, ensuring that recorded frames reflect physically consistent temporal evolution.

\noindent\textbf{Lighting.}
The Render Agent configures illumination to satisfy the descriptive requirements of a scene while maintaining visual and physical coherence. Depending on the rendering backend, it may adopt directional lights, environment maps, or local light sources. The agent adjusts lighting placement and intensity to achieve the requested mood or emphasis, while avoiding artifacts such as visible light emitters or inconsistent illumination. This allows the generated scenes to reflect the intended lighting conditions in a stable and interpretable manner.
\section{Experiments}

\begin{figure*}[t]
    \centering
    \includegraphics[width=0.87\linewidth]{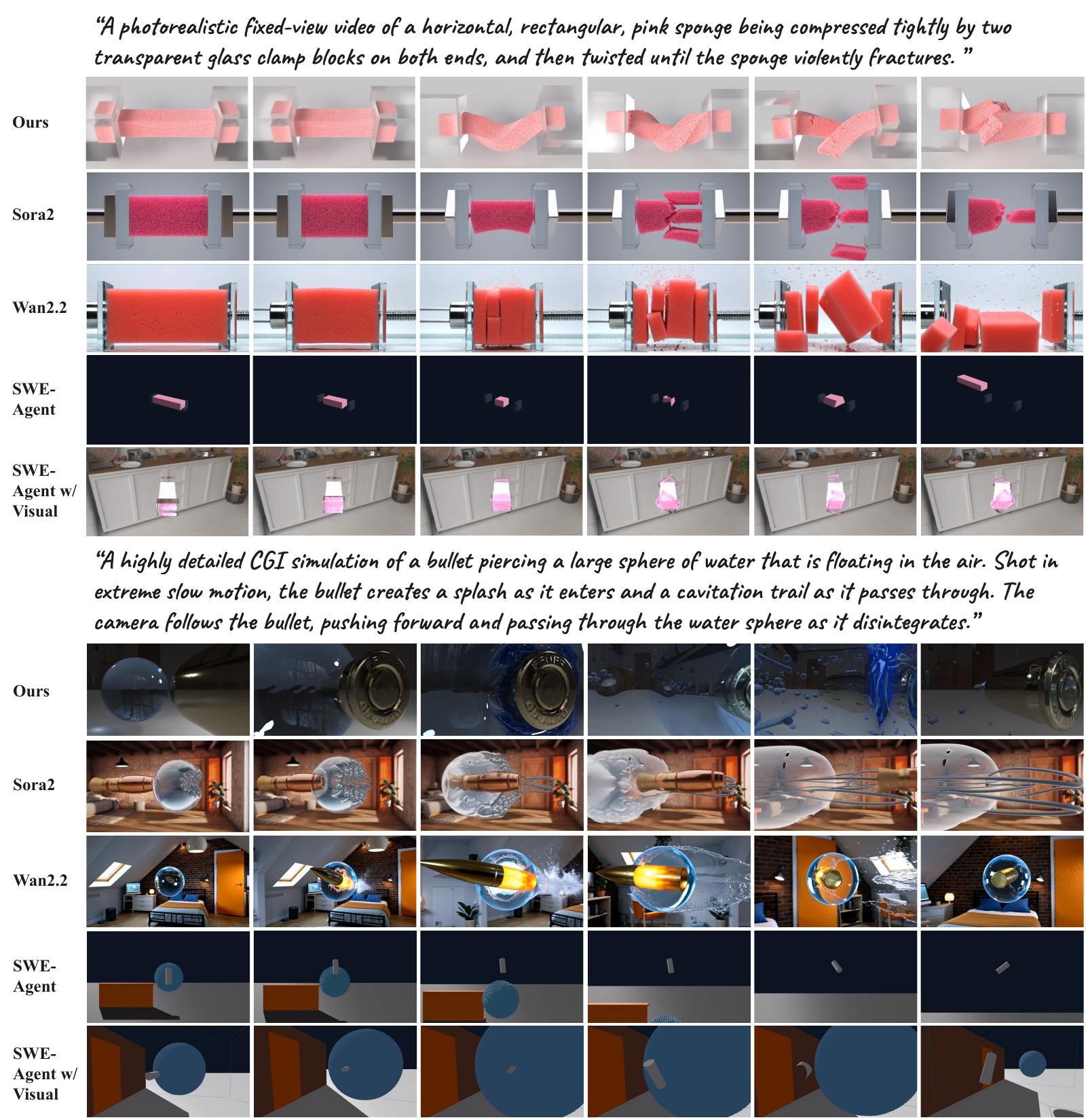}
    \caption{\textbf{Qualitative comparisons} between GS-Agent and the baseline methods. More cases are presented in the Appendix~\ref{app:exp}.}
    \label{fig:gallery}
    \vspace{-4mm}
\end{figure*}

\subsection{Experimental Setup}

To systematically evaluate GS-Agent's ability to construct 4D physical worlds, we conduct experiments across two evaluation suites. The first comprises 24 scenes from the NewtonGen benchmark~\cite{yuan2025newtongen}, specifically designed to test 12 distinct physical laws (e.g., parabolic motion, deformation). To assess the novel capabilities unlocked by our method, we collected a second suite of 30 complex scenes featuring non-continuous dynamics in multi-object interactions (e.g., collisions) and dynamic camera controls.

\noindent\textbf{Metrics.} The NewtonGen benchmark reports the Physical Invariance Score (PIS), which measures the relative standard deviation of theoretically invariant physical quantities (e.g., vertical acceleration during parabolic motion). While the original paper utilizes SAM2 to segment objects per frame and approximate centroids for this calculation (\textbf{Video-PIS}), our agent-simulation framework uniquely unlocks a more rigorous metric. We directly extract exact 3D center-of-mass kinematics from the physics engine at every timestep to compute \textbf{State-PIS}. This ground-truth measurement is inherently inaccessible to standard text-to-video models that only output pixel data. We also measure the \textbf{Alignment Score} between the generated video and text instructions using Perception Encoder feature similarity~\citep{bolya2025perception}, and we adopt the \textbf{Aesthetic} metric from VBench~\citep{huang2024vbench}.

\noindent\textbf{Baselines.} We compare our method against two lines of baselines: text-to-video generative models and foundation agents that can code to solve tasks. For text-to-video generative models, we compare with \textit{Sora-2}~\citep{sora2} and \textit{Wan2.2}~\citep{wan2.2}, representative closed-source and open-source state-of-the-art. For foundation agents, we compare with \textit{SWE-Agent}~\citep{yang2024swe}, a well-established agentic framework for software engineering tasks. Because SWE-Agent natively lacks visual capabilities, we introduce \textit{SWE-Agent w/ Visual}, a variant equipped with a vision-language model (VLM) tool to provide visual feedback on generated videos.

\noindent\textbf{Implementation Details.} We employ Genesis~\citep{genesis} as our underlying physics engine, leveraging its capacity to simulate diverse materials and its fast, photo-realistic rendering system. Unless otherwise specified, we use \texttt{gpt-5} as the foundation model backbone for both our method and the baselines. To ensure a fair comparison, all output is rendered at 720p resolution to match the constraints of the baseline video generative models, although GS-Agent inherently supports rendering at much higher resolutions.

\subsection{Main Results}

\begin{table}[t]
\centering
\caption{\textbf{Main results.} We report the average PIS over 24 NewtonGen scenes and alignment score and aesthetic metric over 30 scenes here, the best is in \textbf{bold}, and the second best is in \underline{underline}.}
\resizebox{0.7\linewidth}{!}{
\begin{tabular}{lcccc}
\toprule
Method  & \makecell{Video-PIS} & \makecell{State-PIS} & \makecell{Alignment\\Score} & Aesthetic \\
\midrule
Sora2~\citep{sora2}             & \underline{0.62} & - & \underline{30.6} & \underline{48.5} \\
Wan2.2~\citep{wan2.2}           & 0.46 & - & 29.8 & \textbf{58.1} \\
SWE-Agent~\citep{yang2024swe}   & 0.41 & 0.44 & 25.8 & 42.0 \\
SWE-Agent w/ Visual             & 0.49 & \underline{0.57} & 26.8 & 44.6 \\
GS-Agent (Ours)                 & \textbf{0.71} & \textbf{0.83} & \textbf{32.2} & 47.6 \\
\bottomrule
\end{tabular}
}
\label{tab:main}
\vspace{-6mm}
\end{table}

\begin{table}[t]
\centering
\caption{\textbf{User study results.} We report the mean score of each method over 270 effective responses.}
\resizebox{0.71\linewidth}{!}{
\begin{tabular}{lcccc}
\toprule
Method      & \makecell{Physical\\Plausibility} & \makecell{Camera\\Controllability} & \makecell{Content\\Alignment} & Aesthetics \\
\midrule
Sora2~\citep{sora2}     & \underline{4.01} & \underline{3.92} & \underline{4.65} & \textbf{4.18} \\
Wan2.2~\citep{wan2.2}   & 2.72 & 3.49 & 4.45 & 3.71 \\
SWE-Agent~\citep{yang2024swe}               & 3.18 & 3.49 & 3.40 & 2.40 \\
SWE-Agent w/ Visual     & 3.25 & 3.96 & 3.72 & 2.44 \\
GS-Agent(Ours)          & \textbf{4.33} & \textbf{4.32} & \textbf{4.70} & \underline{3.86} \\
\bottomrule
\end{tabular}
}
\label{tab:user_study}
\vspace{-5mm}
\end{table}

\noindent\textbf{Comparisons with baselines.} Qualitative comparisons are shown in Figure~\ref{fig:gallery}, with quantitative results detailed in Table~\ref{tab:main}. Overall, GS-Agent produces videos that not only align more faithfully with natural language prompts but also exhibit superior physical plausibility. Text-to-video diffusion models frequently follow textual instructions superficially while violating fundamental physical principles. For instance, in the \textit{sponge fracture} scenario, these models hallucinate abrupt, unrealistic breaking mechanics. Because their generation process is driven purely by text-conditioned pixel distributions, they yield visually appealing frames but fail to maintain temporal consistency or adhere to physical laws.  In the \textit{bullet piercing water sphere} case, Sora2 produces a literal but physically incorrect “cavitation trail,” whereas Wan2.2 generates a video where the bullet passes through the sphere without any realistic fluid disintegration. Additionally, inconsistencies such as a static background bed remaining identical after a 180° camera rotation highlight the lack of true 3D scene reasoning. 
Meanwhile, SWE-Agent and its variant generate physically plausible results but often fail to fully satisfy textual instructions, revealing the difficulty of jointly composing simulation code with accurate material parameters and high-quality rendering. By contrast, GS-Agent successfully balances physical realism, instruction fidelity, and visual coherence, validating our physics-in-the-loop, multi-agent design.

\noindent\textbf{Human evaluators prefer GS-Agent.} Automated evaluation of physical plausibility~\citep{motamed2025travl, zhang2026physion}, camera controllability~\citep{lin2025towards}, and content alignment remains an open research challenge. To provide a more robust assessment, we conducted a user study with 15 participants. Subjects are instructed to rate five randomly ordered videos generated by our method and baselines in a $5$-point Likert Scale, for the aspects of \textit{physical plausibility}, \textit{camera controllability}, \textit{Content Alignment}, and \textit{aesthetics}. Based on $270$ valid responses (Table~\ref{tab:user_study}), human evaluators strongly preferred videos generated by GS-Agent for physical plausibility, camera controllability, and content alignment, though our method slightly lags behind the state-of-the-art closed-source model in pure aesthetics. Further details regarding the study protocol and interface are provided in Appendix~\ref{app:human}.

\subsection{Emergent Capabilities}

\begin{figure}[t]
    \centering
    \includegraphics[width=0.8\linewidth]{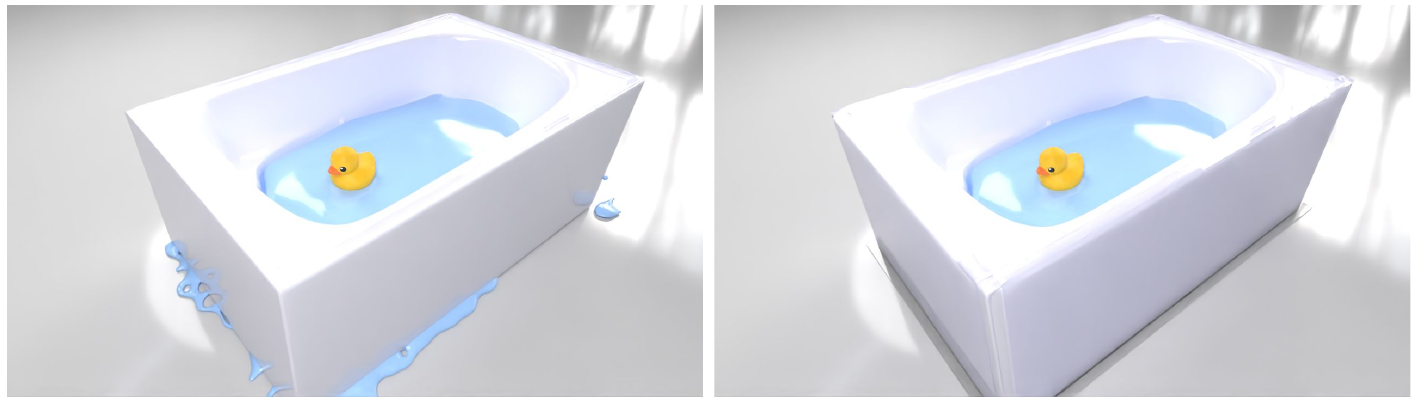}
    \caption{\textbf{Autonomous error recovery.} GS-Agent automatically detects a leaking 3D asset and applies rigid material patches to the corners without human intervention.}
    \label{fig:fix_tub}
\end{figure}

\begin{figure}[t]
    \centering
    \begin{minipage}{0.48\linewidth}
        \centering
        \includegraphics[width=\linewidth]{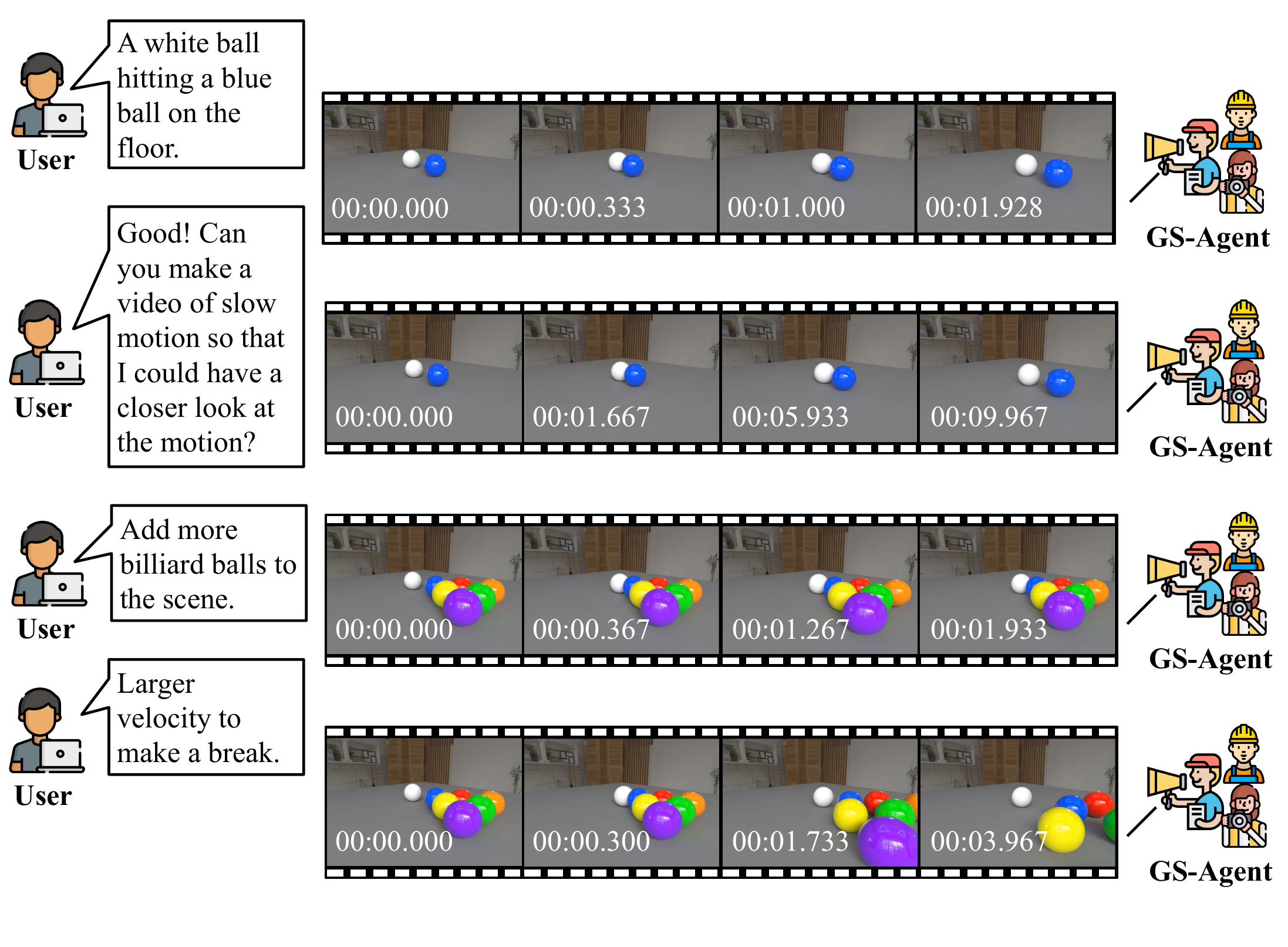}
        \caption{\textbf{Fine-grained controllability.} Users can interact with GS-Agent via natural language to adjust rendering effects (\textit{e.g., slow motion}), create entities (\textit{e.g., add more billiard balls}), and control entity motions (\textit{e.g., increase velocity for a break}).}
        \label{fig:control}
    \end{minipage}
    \hfill
    \begin{minipage}{0.48\linewidth}
        \centering
        \includegraphics[width=\linewidth]{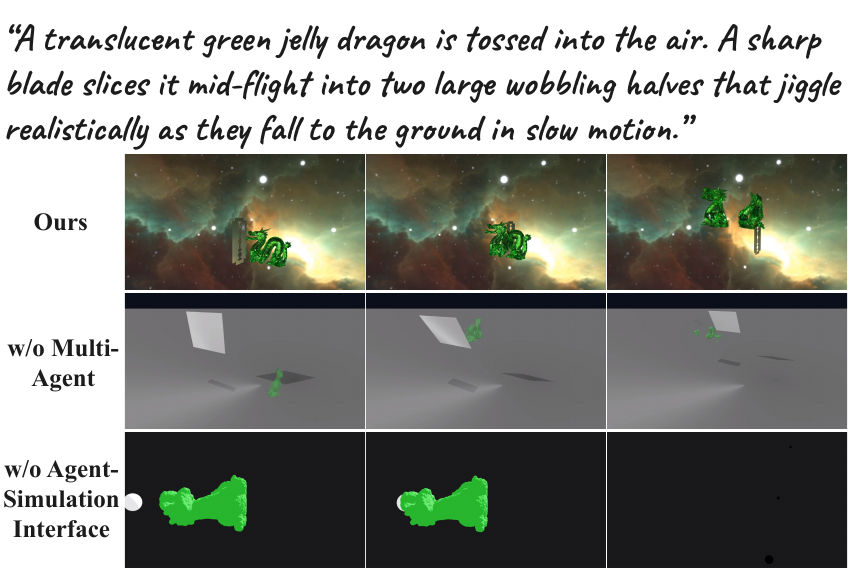}
        \caption{\textbf{Ablation of framework components.} Removing the multi-agent design leads to incomplete scenes, while removing the specialized agent-simulator interface causes the model to struggle with generating stable simulation code.}
        \label{fig:ablate}
    \end{minipage}
    \vspace{-5mm}
\end{figure}

\paragraph{Autonomous error detection and recovery.} Thanks to its flexible end-to-end agentic design, GS-Agent can autonomously detect and resolve unexpected generation failures that would typically require human intervention. Handcrafting robust rules for such edge cases is largely impractical. As illustrated in Figure~\ref{fig:fix_tub}, when an imported 3D bathtub asset was not perfectly waterproof, causing water leakage, GS-Agent recognized the physics failure, proposed a programmatic fix by applying rigid material patches to the geometry's corners, and automatically verified the corrected output. This highlights the system’s capacity to intelligently debug and adapt to unforeseen simulation states.

    

\paragraph{Fine-grained controllability.} As demonstrated in Figure~\ref{fig:control}, users can intuitively manipulate the 4D worlds generated by GS-Agent using natural language. Commands can dictate camera trajectories, spawn new entities, or precisely tune physical parameters like velocity. This level of fine-grained control significantly lowers the barrier to complex content creation and facilitates dynamic visual storytelling.

\begin{table}[t]
    \centering
    \begin{minipage}{0.48\linewidth}
        \centering
        \caption{\textbf{Performance with different backbones.} GS-Agent performs well with various backbones.}
        \label{tab:backbones} 
        \resizebox{\linewidth}{!}{%
        \begin{tabular}{lccc}
        \toprule
        Model & \makecell{State\\PIS} & \makecell{Alignment\\Score} & Aesthetic \\
        \midrule
        \texttt{gpt-5}         & 0.83    & 29.9 & 47.6 \\
        \texttt{gemini-3-pro}  & 0.81    & 28.2 & 45.7 \\
        \texttt{Qwen3.5-27B}   & 0.62    & 27.1 & 45.3 \\
        \bottomrule
        \end{tabular}%
        }
    \end{minipage}
    \hfill 
    \begin{minipage}{0.48\linewidth}
        \centering
        \caption{\textbf{Ablation study results.}}
        \label{tab:ablate} 
        \resizebox{\linewidth}{!}{%
        \begin{tabular}{lccc}
        \toprule
        Method & \makecell{State\\PIS} & \makecell{Alignment\\Score} & Aesthetic \\
        \midrule
        GS-Agent                                      & 0.83    & 29.9 & 47.6 \\
        w/o Multi-Agent                               & 0.42 & 27.9 & 47.3 \\
        \makecell[l]{w/o Agent-Sim Interface} & 0.57 & 26.8 & 44.6 \\
        \bottomrule
        \end{tabular}%
        }
    \end{minipage}
\end{table}

\subsection{Ablation and Backbone Analysis}

\paragraph{The efficacy of multi-agent design.}
We ablate our proposed architecture by consolidating all tool access into a single monolithic agent. As shown in Table~\ref{tab:ablate} and Figure~\ref{fig:ablate}, the multi-agent design achieves markedly superior performance by decomposing the world-generation pipeline into specialized sub-tasks. Confining each agent to a well-defined context reduces prompt distraction and minimizes tool misuse, leading to more reliable reasoning and execution. Consistent with findings in~\cite{li2023camel}, multi-agent systems effectively expand the usable context window for logic reasoning, unlocking advanced task-solving capabilities.

\paragraph{The importance of a specialized Agent-Simulator Interface.}
To isolate the impact of our interface, we compare GS-Agent against the \textit{SWE-Agent w/ Visual} baseline. Generic physics simulator APIs are traditionally designed for human experts; exposing these vast, low-level action spaces directly to autonomous agents frequently results in syntax errors, catastrophic parameter choices, or physics engine crashes. By abstracting raw simulator operations into structured, semantically meaningful tools, our interface constrains agent behavior while preserving creative expressiveness, directly resulting in stable and physically accurate world generation.

\paragraph{Robustness Across LLM Backbones.} We evaluated the framework utilizing \texttt{gemini-3-pro} and the open-weights \texttt{Qwen3.5-27B} models. As detailed in Table~\ref{tab:backbones}, GS-Agent successfully operates across diverse backbones. While the smaller \texttt{Qwen3.5-27B} incurs an expected performance penalty largely due to the rigorous spatial reasoning and dynamic video comprehension required, the overall pipeline remains intact. We anticipate framework performance will scale naturally alongside the rapid evolution of open-weight foundation models.
\section{Discussions}

\textbf{GS-Agent generates more than videos.} While our quantitative metrics focus on rendered video output, \textbf{GS-Agent} fundamentally constructs a multimodal 4D environment. It naturally yields rich, aligned data structures including metric depth maps, precise segmentation masks, surface normals, and particle-level dynamics. Because the final output is an executable simulation script, it can be seamlessly integrated with robotic embodiments for downstream reinforcement learning and evaluation. Consequently, GS-Agent serves not merely as a video generator, but as an engine for embodied data synthesis.

\textbf{Guaranteed physical and temporal consistency.} Maintaining spatiotemporal consistency is a critical bottleneck in developing interactive world models like \textit{Genie3}~\citep{deepmind_genie3_2025}. While pure diffusion models struggle to preserve geometric coherence over long horizons~\citep{xiang2024pandora}, our methodology grounds world construction in procedural code executed by a deterministic physics engine. This approach mathematically guarantees the internal consistency of geometry, motion, and material interactions, offering a robust foundation for future controllable world models.

\subsection{Limitations}
\label{sec:limitation}

\textbf{Constraints of Current Simulation Technology.}
Accurately and efficiently simulating the real world remains an open challenge~\citep{boeing2007evaluation}. The fidelity and scalability of contemporary graphics and physics simulation algorithms inherently constrain the performance ceiling of our method. Nevertheless, even with current simulation tools, humans can already craft highly complex 4D worlds, such as those in modern games and films, albeit with extensive manual effort. Our work aims to automate this process and leverage continual advancements in graphics and simulation~\citep{chen2025offset}, advocating a promising path toward scalable, physically grounded world generation.

\textbf{Limitations in Foundation Model Cinematic Understanding.} Our method relies on foundation models’ capacity for self-critique, particularly their ability to evaluate whether a generated scene exhibits expected motion and coherent cinematic expression. While advanced models such as GPT-5 have demonstrated notable improvements in these domains, they remain far from perfect~\citep{kang2024far, liu2025shotbench}. Continued progress in motion and cinematic understanding~\citep{motamed2025travl,lin2025towards} will further enhance our framework’s ability to produce fine-grained, physically consistent, and aesthetically rich generations.
\section{Conclusion}

In this work, we presented GS-Agent, a multi-agent framework that integrates physics engine in the loop to generate dynamic and physically plausible 4D worlds from natural languages. GS-Agent interacts with the physics engine through executable code, actively seeks multi-modal feedback, and collaborates with others to iteratively construct coherent and controllable worlds. Experimental results demonstrate that GS-Agent produces physically plausible worlds and offers fine-grained controllability beyond existing methods. Looking forward, we envision our approach as a foundation for building large-scale, interactive, and consistent world simulators, bridging the gap between generative models, embodied intelligence, and physics-based simulation for scalable multimodal data creation and new formats of experience.



\newpage
\newpage

\clearpage
\bibliography{main}

@String(CVPR= {IEEE Conf. Comput. Vis. Pattern Recog.})

@String(TOG= {ACM Trans. Graph.})

@String(CVPR  = {CVPR})

@String(TOG   = {ACM TOG})

@inproceedings{chang2014learning,
  title={Learning spatial knowledge for text to 3D scene generation},
  author={Chang, Angel and Savva, Manolis and Manning, Christopher D},
  booktitle={Proceedings of the 2014 conference on empirical methods in natural language processing (EMNLP)},
  pages={2028--2038},
  year={2014}
}

@article{yang2024swe,
  title={Swe-agent: Agent-computer interfaces enable automated software engineering},
  author={Yang, John and Jimenez, Carlos E and Wettig, Alexander and Lieret, Kilian and Yao, Shunyu and Narasimhan, Karthik and Press, Ofir},
  journal={Advances in Neural Information Processing Systems},
  volume={37},
  pages={50528--50652},
  year={2024}
}

@inproceedings{hu2024scenecraft,
  title={Scenecraft: An llm agent for synthesizing 3d scenes as blender code},
  author={Hu, Ziniu and Iscen, Ahmet and Jain, Aashi and Kipf, Thomas and Yue, Yisong and Ross, David A and Schmid, Cordelia and Fathi, Alireza},
  booktitle={Forty-first International Conference on Machine Learning},
  year={2024}
}

@inproceedings{sun20253d,
  title={3d-gpt: Procedural 3d modeling with large language models},
  author={Sun, Chunyi and Han, Junlin and Deng, Weijian and Wang, Xinlong and Qin, Zishan and Gould, Stephen},
  booktitle={2025 International Conference on 3D Vision (3DV)},
  pages={1253--1263},
  year={2025},
  organization={IEEE}
}

@inproceedings{lv2024gpt4motion,
  title={Gpt4motion: Scripting physical motions in text-to-video generation via blender-oriented gpt planning},
  author={Lv, Jiaxi and Huang, Yi and Yan, Mingfu and Huang, Jiancheng and Liu, Jianzhuang and Liu, Yifan and Wen, Yafei and Chen, Xiaoxin and Chen, Shifeng},
  booktitle={Proceedings of the IEEE/CVF conference on computer vision and pattern recognition},
  pages={1430--1440},
  year={2024}
}

@article{yuan2025newtongen,
  title={NewtonGen: Physics-Consistent and Controllable Text-to-Video Generation via Neural Newtonian Dynamics},
  author={Yuan, Yu and Wang, Xijun and Wickremasinghe, Tharindu and Nadir, Zeeshan and Ma, Bole and Chan, Stanley H},
  journal={arXiv preprint arXiv:2509.21309},
  year={2025}
}

@inproceedings{bruce2024genie,
  title={Genie: Generative interactive environments},
  author={Bruce, Jake and Dennis, Michael D and Edwards, Ashley and Parker-Holder, Jack and Shi, Yuge and Hughes, Edward and Lai, Matthew and Mavalankar, Aditi and Steigerwald, Richie and Apps, Chris and others},
  booktitle={Forty-first International Conference on Machine Learning},
  year={2024}
}

@article{bah20244dfy,
  author = {Bahmani, Sherwin and Skorokhodov, Ivan and Rong, Victor and Wetzstein, Gordon and Guibas, Leonidas and Wonka, Peter and Tulyakov, Sergey and Park, Jeong Joon and Tagliasacchi, Andrea and Lindell, David B.},
  title = {4D-fy: Text-to-4D Generation Using Hybrid Score Distillation Sampling},
  journal = {IEEE Conference on Computer Vision and Pattern Recognition ({CVPR})},
  year = {2024},
}

@article{xu2024comp4d,
  title={Comp4d: Llm-guided compositional 4d scene generation},
  author={Xu, Dejia and Liang, Hanwen and Bhatt, Neel P and Hu, Hezhen and Liang, Hanxue and Plataniotis, Konstantinos N and Wang, Zhangyang},
  journal={arXiv preprint arXiv:2403.16993},
  year={2024}
}

@article{meng2024towards,
  title={Towards world simulator: Crafting physical commonsense-based benchmark for video generation},
  author={Meng, Fanqing and Liao, Jiaqi and Tan, Xinyu and Shao, Wenqi and Lu, Quanfeng and Zhang, Kaipeng and Cheng, Yu and Li, Dianqi and Qiao, Yu and Luo, Ping},
  journal={arXiv preprint arXiv:2410.05363},
  year={2024}
}

@article{motamed2025travl,
  title={TRAVL: A Recipe for Making Video-Language Models Better Judges of Physics Implausibility},
  author={Motamed, Saman and Chen, Minghao and Van Gool, Luc and Laina, Iro},
  journal={arXiv preprint arXiv:2510.07550},
  year={2025}
}

@article{he2024cameractrl,
  title={Cameractrl: Enabling camera control for text-to-video generation},
  author={He, Hao and Xu, Yinghao and Guo, Yuwei and Wetzstein, Gordon and Dai, Bo and Li, Hongsheng and Yang, Ceyuan},
  journal={arXiv preprint arXiv:2404.02101},
  year={2024}
}

@inproceedings{yu2024wonderjourney,
  title={Wonderjourney: Going from anywhere to everywhere},
  author={Yu, Hong-Xing and Duan, Haoyi and Hur, Junhwa and Sargent, Kyle and Rubinstein, Michael and Freeman, William T and Cole, Forrester and Sun, Deqing and Snavely, Noah and Wu, Jiajun and others},
  booktitle={Proceedings of the IEEE/CVF Conference on Computer Vision and Pattern Recognition},
  pages={6658--6667},
  year={2024}
}

@inproceedings{yu2025wonderworld,
  title={Wonderworld: Interactive 3d scene generation from a single image},
  author={Yu, Hong-Xing and Duan, Haoyi and Herrmann, Charles and Freeman, William T and Wu, Jiajun},
  booktitle={Proceedings of the Computer Vision and Pattern Recognition Conference},
  pages={5916--5926},
  year={2025}
}

@article{li2025wonderplay,
  title={WonderPlay: Dynamic 3D Scene Generation from a Single Image and Actions},
  author={Li, Zizhang and Yu, Hong-Xing and Liu, Wei and Yang, Yin and Herrmann, Charles and Wetzstein, Gordon and Wu, Jiajun},
  journal={arXiv preprint arXiv:2505.18151},
  year={2025}
}

@inproceedings{chen2025physgen3d,
  title={Physgen3d: Crafting a miniature interactive world from a single image},
  author={Chen, Boyuan and Jiang, Hanxiao and Liu, Shaowei and Gupta, Saurabh and Li, Yunzhu and Zhao, Hao and Wang, Shenlong},
  booktitle={Proceedings of the Computer Vision and Pattern Recognition Conference},
  pages={6178--6189},
  year={2025}
}

@article{chung2023luciddreamer,
  title={Luciddreamer: Domain-free generation of 3d gaussian splatting scenes},
  author={Chung, Jaeyoung and Lee, Suyoung and Nam, Hyeongjin and Lee, Jaerin and Lee, Kyoung Mu},
  journal={arXiv preprint arXiv:2311.13384},
  year={2023}
}

@inproceedings{huang2024vbench,
  title={Vbench: Comprehensive benchmark suite for video generative models},
  author={Huang, Ziqi and He, Yinan and Yu, Jiashuo and Zhang, Fan and Si, Chenyang and Jiang, Yuming and Zhang, Yuanhan and Wu, Tianxing and Jin, Qingyang and Chanpaisit, Nattapol and others},
  booktitle={Proceedings of the IEEE/CVF Conference on Computer Vision and Pattern Recognition},
  pages={21807--21818},
  year={2024}
}

@inproceedings{yangcogvideox,
  title={CogVideoX: Text-to-Video Diffusion Models with An Expert Transformer},
  author={Yang, Zhuoyi and Teng, Jiayan and Zheng, Wendi and Ding, Ming and Huang, Shiyu and Xu, Jiazheng and Yang, Yuanming and Hong, Wenyi and Zhang, Xiaohan and Feng, Guanyu and others},
  booktitle={The Thirteenth International Conference on Learning Representations},
  year = {2024}
}

@inproceedings{radford2021learning,
  title={Learning transferable visual models from natural language supervision},
  author={Radford, Alec and Kim, Jong Wook and Hallacy, Chris and Ramesh, Aditya and Goh, Gabriel and Agarwal, Sandhini and Sastry, Girish and Askell, Amanda and Mishkin, Pamela and Clark, Jack and others},
  booktitle={International conference on machine learning},
  pages={8748--8763},
  year={2021},
  organization={PmLR}
}

@article{wan2.2,
      title={Wan: Open and Advanced Large-Scale Video Generative Models}, 
      author={Team Wan and Ang Wang and Baole Ai and Bin Wen and Chaojie Mao and Chen-Wei Xie and Di Chen and Feiwu Yu and Haiming Zhao and Jianxiao Yang and Jianyuan Zeng and Jiayu Wang and Jingfeng Zhang and Jingren Zhou and Jinkai Wang and Jixuan Chen and Kai Zhu and Kang Zhao and Keyu Yan and Lianghua Huang and Mengyang Feng and Ningyi Zhang and Pandeng Li and Pingyu Wu and Ruihang Chu and Ruili Feng and Shiwei Zhang and Siyang Sun and Tao Fang and Tianxing Wang and Tianyi Gui and Tingyu Weng and Tong Shen and Wei Lin and Wei Wang and Wei Wang and Wenmeng Zhou and Wente Wang and Wenting Shen and Wenyuan Yu and Xianzhong Shi and Xiaoming Huang and Xin Xu and Yan Kou and Yangyu Lv and Yifei Li and Yijing Liu and Yiming Wang and Yingya Zhang and Yitong Huang and Yong Li and You Wu and Yu Liu and Yulin Pan and Yun Zheng and Yuntao Hong and Yupeng Shi and Yutong Feng and Zeyinzi Jiang and Zhen Han and Zhi-Fan Wu and Ziyu Liu},
      journal = {arXiv preprint arXiv:2503.20314},
      year={2025}
}

@software{sora2,
  title        = {Sora 2},
  author       = {{OpenAI}},
  year         = {2025},
  url          = {https://openai.com/index/sora-2/},
  note         = {Video and audio generation model; released 30 Sep 2025}
}

@misc{Genesis,
          author = {Genesis Authors},
          title = {Genesis: A Generative and Universal Physics Engine for Robotics and Beyond},
          month = {December},
          year = {2024},
          url = {https://github.com/Genesis-Embodied-AI/Genesis}
        }

@article{ho2022imagen,
  title={Imagen video: High definition video generation with diffusion models},
  author={Ho, Jonathan and Chan, William and Saharia, Chitwan and Whang, Jay and Gao, Ruiqi and Gritsenko, Alexey and Kingma, Diederik P and Poole, Ben and Norouzi, Mohammad and Fleet, David J and others},
  journal={arXiv preprint arXiv:2210.02303},
  year={2022}
}

@article{blattmann2023stable,
  title={Stable video diffusion: Scaling latent video diffusion models to large datasets},
  author={Blattmann, Andreas and Dockhorn, Tim and Kulal, Sumith and Mendelevitch, Daniel and Kilian, Maciej and Lorenz, Dominik and Levi, Yam and English, Zion and Voleti, Vikram and Letts, Adam and others},
  journal={arXiv preprint arXiv:2311.15127},
  year={2023}
}

@misc{openai2023gpt4,
      title={GPT-4 Technical Report}, 
      author={OpenAI},
      year={2023},
      eprint={2303.08774},
      archivePrefix={arXiv},
      primaryClass={cs.CL}
}

@article{liu2023visual,
  title={Visual instruction tuning},
  author={Liu, Haotian and Li, Chunyuan and Wu, Qingyang and Lee, Yong Jae},
  journal={arXiv preprint arXiv:2304.08485},
  year={2023}
}

@article{ho2020denoising,
  title={Denoising diffusion probabilistic models},
  author={Ho, Jonathan and Jain, Ajay and Abbeel, Pieter},
  journal={Advances in neural information processing systems},
  volume={33},
  pages={6840--6851},
  year={2020}
}

@article{guo2025deepseek,
  title={DeepSeek-R1: Incentivizing Reasoning Capability in LLMs via Reinforcement Learning},
  author={Guo, Daya and Yang, Dejian and Zhang, Haowei and Song, Junxiao and Zhang, Ruoyu and Xu, Runxin and Zhu, Qihao and Ma, Shirong and Wang, Peiyi and Bi, Xiao and others},
  journal={arXiv preprint arXiv:2501.12948},
  year={2025}
}

@article{kang2024far,
  title={How far is video generation from world model: A physical law perspective},
  author={Kang, Bingyi and Yue, Yang and Lu, Rui and Lin, Zhijie and Zhao, Yang and Wang, Kaixin and Huang, Gao and Feng, Jiashi},
  journal={arXiv preprint arXiv:2411.02385},
  year={2024}
}

@misc{meshy2024,
  author = {{Meshy}},
  title = {Meshy: Generative AI for 3D Content Creation},
  year = {2024},
  url = {https://www.meshy.ai/},
  note = {Accessed: 2026-01-29}
}

@inproceedings{yu2023magvit,
  title={Magvit: Masked generative video transformer},
  author={Yu, Lijun and Cheng, Yong and Sohn, Kihyuk and Lezama, Jos{\'e} and Zhang, Han and Chang, Huiwen and Hauptmann, Alexander G and Yang, Ming-Hsuan and Hao, Yuan and Essa, Irfan and others},
  booktitle={Proceedings of the IEEE/CVF Conference on Computer Vision and Pattern Recognition},
  pages={10459--10469},
  year={2023}
}

@article{kondratyuk2023videopoet,
  title={Videopoet: A large language model for zero-shot video generation},
  author={Kondratyuk, Dan and Yu, Lijun and Gu, Xiuye and Lezama, Jos{\'e} and Huang, Jonathan and Schindler, Grant and Hornung, Rachel and Birodkar, Vighnesh and Yan, Jimmy and Chiu, Ming-Chang and others},
  journal={arXiv preprint arXiv:2312.14125},
  year={2023}
}

@inproceedings{peebles2023scalable,
  title={Scalable diffusion models with transformers},
  author={Peebles, William and Xie, Saining},
  booktitle={Proceedings of the IEEE/CVF international conference on computer vision},
  pages={4195--4205},
  year={2023}
}

@inproceedings{chen2024videocrafter2,
  title={Videocrafter2: Overcoming data limitations for high-quality video diffusion models},
  author={Chen, Haoxin and Zhang, Yong and Cun, Xiaodong and Xia, Menghan and Wang, Xintao and Weng, Chao and Shan, Ying},
  booktitle={Proceedings of the IEEE/CVF Conference on Computer Vision and Pattern Recognition},
  pages={7310--7320},
  year={2024}
}

@InProceedings{Xie_2024_CVPR,
    author    = {Xie, Tianyi and Zong, Zeshun and Qiu, Yuxing and Li, Xuan and Feng, Yutao and Yang, Yin and Jiang, Chenfanfu},
    title     = {PhysGaussian: Physics-Integrated 3D Gaussians for Generative Dynamics},
    booktitle = {Proceedings of the IEEE/CVF Conference on Computer Vision and Pattern Recognition (CVPR)},
    month     = {June},
    year      = {2024},
    pages     = {4389-4398}
}

@InProceedings{Zhang_2025_CVPR,
    author    = {Zhang, Zhenghao and Liao, Junchao and Li, Menghao and Dai, ZuoZhuo and Qiu, Bingxue and Zhu, Siyu and Qin, Long and Wang, Weizhi},
    title     = {Tora: Trajectory-oriented Diffusion Transformer for Video Generation},
    booktitle = {Proceedings of the IEEE/CVF Conference on Computer Vision and Pattern Recognition (CVPR)},
    month     = {June},
    year      = {2025},
    pages     = {2063-2073}
}

@inproceedings{sohl2015deep,
  title={Deep unsupervised learning using nonequilibrium thermodynamics},
  author={Sohl-Dickstein, Jascha and Weiss, Eric and Maheswaranathan, Niru and Ganguli, Surya},
  booktitle={International conference on machine learning},
  pages={2256--2265},
  year={2015},
  organization={pmlr}
}

@inproceedings{liu2024physgen,
  title={Physgen: Rigid-body physics-grounded image-to-video generation},
  author={Liu, Shaowei and Ren, Zhongzheng and Gupta, Saurabh and Wang, Shenlong},
  booktitle={European Conference on Computer Vision},
  pages={360--378},
  year={2024},
  organization={Springer}
}

@article{ma2018language,
  title={Language-driven synthesis of 3D scenes from scene databases},
  author={Ma, Rui and Patil, Akshay Gadi and Fisher, Matthew and Li, Manyi and Pirk, S{\"o}ren and Hua, Binh-Son and Yeung, Sai-Kit and Tong, Xin and Guibas, Leonidas and Zhang, Hao},
  journal={ACM Transactions on Graphics (TOG)},
  volume={37},
  number={6},
  pages={1--16},
  year={2018},
  publisher={ACM New York, NY, USA}
}

@article{chang2015text,
  title={Text to 3d scene generation with rich lexical grounding},
  author={Chang, Angel and Monroe, Will and Savva, Manolis and Potts, Christopher and Manning, Christopher D},
  journal={arXiv preprint arXiv:1505.06289},
  year={2015}
}

@inproceedings{zhang2024physdreamer,
    title={{PhysDreamer}: Physics-Based Interaction with 3D Objects via Video Generation},
    author={Tianyuan Zhang and Hong-Xing Yu and Rundi Wu and Brandon Y. Feng and Changxi Zheng and Noah Snavely and Jiajun Wu and William T. Freeman},
    booktitle={European Conference on Computer Vision},
    year={2024},
    organization={Springer}
    }

@inproceedings{zhang2024building,
  title={Building Cooperative Embodied Agents Modularly with Large Language Models},
  author={Zhang, Hongxin and Du, Weihua and Shan, Jiaming and Zhou, Qinhong and Du, Yilun and Tenenbaum, Joshua B and Shu, Tianmin and Gan, Chuang},
  booktitle={The Twelfth International Conference on Learning Representations},
  year={2024}
}

@inproceedings{du2023improving,
  title={Improving factuality and reasoning in language models through multiagent debate},
  author={Du, Yilun and Li, Shuang and Torralba, Antonio and Tenenbaum, Joshua B and Mordatch, Igor},
  booktitle={Forty-first International Conference on Machine Learning},
  year={2023}
}

@inproceedings{
liu2024a,
title={A Dynamic {LLM}-Powered Agent Network for Task-Oriented Agent Collaboration},
author={Zijun Liu and Yanzhe Zhang and Peng Li and Yang Liu and Diyi Yang},
booktitle={First Conference on Language Modeling},
year={2024},
url={https://openreview.net/forum?id=XII0Wp1XA9}
}

@article{li2023camel,
  title={Camel: Communicative agents for" mind" exploration of large language model society},
  author={Li, Guohao and Hammoud, Hasan and Itani, Hani and Khizbullin, Dmitrii and Ghanem, Bernard},
  journal={Advances in Neural Information Processing Systems},
  volume={36},
  pages={51991--52008},
  year={2023}
}

@inproceedings{wu2024autogen,
  title={Autogen: Enabling next-gen LLM applications via multi-agent conversations},
  author={Wu, Qingyun and Bansal, Gagan and Zhang, Jieyu and Wu, Yiran and Li, Beibin and Zhu, Erkang and Jiang, Li and Zhang, Xiaoyun and Zhang, Shaokun and Liu, Jiale and others},
  booktitle={First Conference on Language Modeling},
  year={2024}
}

@inproceedings{hong2024cogagent,
  title={Cogagent: A visual language model for gui agents},
  author={Hong, Wenyi and Wang, Weihan and Lv, Qingsong and Xu, Jiazheng and Yu, Wenmeng and Ji, Junhui and Wang, Yan and Wang, Zihan and Dong, Yuxiao and Ding, Ming and others},
  booktitle={Proceedings of the IEEE/CVF Conference on Computer Vision and Pattern Recognition},
  pages={14281--14290},
  year={2024}
}

@article{gur2023real,
  title={A real-world webagent with planning, long context understanding, and program synthesis},
  author={Gur, Izzeddin and Furuta, Hiroki and Huang, Austin and Safdari, Mustafa and Matsuo, Yutaka and Eck, Douglas and Faust, Aleksandra},
  journal={arXiv preprint arXiv:2307.12856},
  year={2023}
}

@article{shinn2024reflexion,
  title={Reflexion: Language agents with verbal reinforcement learning},
  author={Shinn, Noah and Cassano, Federico and Gopinath, Ashwin and Narasimhan, Karthik and Yao, Shunyu},
  journal={Advances in Neural Information Processing Systems},
  volume={36},
  year={2024}
}

@article{du2023video,
  title={Video language planning},
  author={Du, Yilun and Yang, Mengjiao and Florence, Pete and Xia, Fei and Wahid, Ayzaan and Ichter, Brian and Sermanet, Pierre and Yu, Tianhe and Abbeel, Pieter and Tenenbaum, Joshua B and others},
  journal={arXiv preprint arXiv:2310.10625},
  year={2023}
}

@article{agashe2025agent,
  title={Agent s2: A compositional generalist-specialist framework for computer use agents},
  author={Agashe, Saaket and Wong, Kyle and Tu, Vincent and Yang, Jiachen and Li, Ang and Wang, Xin Eric},
  journal={arXiv preprint arXiv:2504.00906},
  year={2025}
}

@inproceedings{huang2023voxposer,
  title={VoxPoser: Composable 3D Value Maps for Robotic Manipulation with Language Models},
  author={Huang, Wenlong and Wang, Chen and Zhang, Ruohan and Li, Yunzhu and Wu, Jiajun and Fei-Fei, Li},
  booktitle={Conference on Robot Learning},
  pages={540--562},
  year={2023},
  organization={PMLR}
}

@article{ahn2022can,
  title={Do as i can, not as i say: Grounding language in robotic affordances},
  author={Ahn, Michael and Brohan, Anthony and Brown, Noah and Chebotar, Yevgen and Cortes, Omar and David, Byron and Finn, Chelsea and Gopalakrishnan, Keerthana and Hausman, Karol and Herzog, Alex and others},
  journal={arXiv preprint arXiv:2204.01691},
  year={2022}
}

@article{xi2023rise,
  title={The Rise and Potential of Large Language Model Based Agents: A Survey},
  author={Xi, Zhiheng and Chen, Wenxiang and Guo, Xin and He, Wei and Ding, Yiwen and Hong, Boyang and Zhang, Ming and Wang, Junzhe and Jin, Senjie and Zhou, Enyu and others},
  journal={arXiv preprint arXiv:2309.07864},
  year={2023}
}

@article{wang2023survey,
  title={A survey on large language model based autonomous agents},
  author={Wang, Lei and Ma, Chen and Feng, Xueyang and Zhang, Zeyu and Yang, Hao and Zhang, Jingsen and Chen, Zhiyuan and Tang, Jiakai and Chen, Xu and Lin, Yankai and others},
  journal={arXiv preprint arXiv:2308.11432},
  year={2023}
}

@article{sumers2023cognitive,
  title={Cognitive architectures for language agents},
  author={Sumers, Theodore and Yao, Shunyu and Narasimhan, Karthik and Griffiths, Thomas L},
  journal={arXiv preprint arXiv:2309.02427},
  year={2023}
}

@article{qiu2025alita,
  title={Alita: Generalist agent enabling scalable agentic reasoning with minimal predefinition and maximal self-evolution},
  author={Qiu, Jiahao and Qi, Xuan and Zhang, Tongcheng and Juan, Xinzhe and Guo, Jiacheng and Lu, Yifu and Wang, Yimin and Yao, Zixin and Ren, Qihan and Jiang, Xun and others},
  journal={arXiv preprint arXiv:2505.20286},
  year={2025}
}

@article{hu2025owl,
  title={Owl: Optimized workforce learning for general multi-agent assistance in real-world task automation},
  author={Hu, Mengkang and Zhou, Yuhang and Fan, Wendong and Nie, Yuzhou and Xia, Bowei and Sun, Tao and Ye, Ziyu and Jin, Zhaoxuan and Li, Yingru and Chen, Qiguang and others},
  journal={arXiv preprint arXiv:2505.23885},
  year={2025}
}

@incollection{jiang2016material,
  title={The material point method for simulating continuum materials},
  author={Jiang, Chenfanfu and Schroeder, Craig and Teran, Joseph and Stomakhin, Alexey and Selle, Andrew},
  booktitle={Acm siggraph 2016 courses},
  pages={1--52},
  year={2016}
}

@inproceedings{muller2003particle,
  title={Particle-based fluid simulation for interactive applications},
  author={M{\"u}ller, Matthias and Charypar, David and Gross, Markus},
  booktitle={Proceedings of the 2003 ACM SIGGRAPH/Eurographics symposium on Computer animation},
  pages={154--159},
  year={2003}
}

@inproceedings{liu2012physicalMaterialEditing,
  author    = {Ning Liu and Xue-Wei Hua and Renyi Zhang and Lisheng Wang},
  title     = {Physical Material Editing with Structure Embedding for Realistic Deformable Objects},
  booktitle = {Proceedings of the ACM Symposium on Interactive 3D Graphics and Games (I3D)},
  year      = {2012}
}

@inproceedings{zhao2020sim,
  title={Sim-to-real transfer in deep reinforcement learning for robotics: a survey},
  author={Zhao, Wenshuai and Queralta, Jorge Pe{\~n}a and Westerlund, Tomi},
  booktitle={2020 IEEE symposium series on computational intelligence (SSCI)},
  pages={737--744},
  year={2020},
  organization={IEEE}
}

@article{wang2024architect,
  title={Architect: Generating vivid and interactive 3d scenes with hierarchical 2d inpainting},
  author={Wang, Yian and Qiu, Xiaowen and Liu, Jiageng and Chen, Zhehuan and Cai, Jiting and Wang, Yufei and Wang, Tsun-Hsuan and Xian, Zhou and Gan, Chuang},
  journal={Advances in Neural Information Processing Systems},
  volume={37},
  pages={67575--67603},
  year={2024}
}

@article{lin2025towards,
  title={Towards Understanding Camera Motions in Any Video},
  author={Lin, Zhiqiu and Cen, Siyuan and Jiang, Daniel and Karhade, Jay and Wang, Hewei and Mitra, Chancharik and Ling, Tiffany and Huang, Yuhan and Liu, Sifan and Chen, Mingyu and others},
  journal={arXiv preprint arXiv:2504.15376},
  year={2025}
}

@article{xiang2024pandora,
  title={Pandora: Towards general world model with natural language actions and video states},
  author={Xiang, Jiannan and Liu, Guangyi and Gu, Yi and Gao, Qiyue and Ning, Yuting and Zha, Yuheng and Feng, Zeyu and Tao, Tianhua and Hao, Shibo and Shi, Yemin and others},
  journal={arXiv preprint arXiv:2406.09455},
  year={2024}
}

@misc{deepmind_genie3_2025,
  author       = {Parker-Holder, Jack and Fruchter, Shlomi},
  title        = {Genie 3: A New Frontier for World Models},
  year         = {2025},
  howpublished = {Google DeepMind Blog},
  url          = {https://deepmind.google/blog/genie-3-a-new-frontier-for-world-models/}
}

@article{liu2025shotbench,
  title={ShotBench: Expert-Level Cinematic Understanding in Vision-Language Models},
  author={Liu, Hongbo and He, Jingwen and Jin, Yi and Zheng, Dian and Dong, Yuhao and Zhang, Fan and Huang, Ziqi and He, Yinan and Li, Yangguang and Chen, Weichao and others},
  journal={arXiv preprint arXiv:2506.21356},
  year={2025}
}

@article{chen2025offset,
  title={Offset Geometric Contact},
  author={Chen, Anka He and Hsu, Jerry and Liu, Ziheng and Macklin, Miles and Yang, Yin and Yuksel, Cem},
  journal={ACM Transactions on Graphics (TOG)},
  volume={44},
  number={4},
  pages={1--21},
  year={2025},
  publisher={ACM New York, NY, USA}
}

@inproceedings{boeing2007evaluation,
  title={Evaluation of real-time physics simulation systems},
  author={Boeing, Adrian and Br{\"a}unl, Thomas},
  booktitle={Proceedings of the 5th international conference on Computer graphics and interactive techniques in Australia and Southeast Asia},
  pages={281--288},
  year={2007}
}

@article{bolya2025perception,
  title={Perception encoder: The best visual embeddings are not at the output of the network},
  author={Bolya, Daniel and Huang, Po-Yao and Sun, Peize and Cho, Jang Hyun and Madotto, Andrea and Wei, Chen and Ma, Tengyu and Zhi, Jiale and Rajasegaran, Jathushan and Rasheed, Hanoona and others},
  journal={arXiv preprint arXiv:2504.13181},
  year={2025}
}

@article{wang2024diffusion,
  title={Diffusion models in 3d vision: A survey},
  author={Wang, Zhen and Li, Dongyuan and Wu, Yaozu and He, Tianyu and Bian, Jiang and Jiang, Renhe},
  journal={arXiv preprint arXiv:2410.04738},
  year={2024}
}

@article{zhang2026physion,
  title={Physion-Eval: Evaluating Physical Realism in Generated Video via Human Reasoning},
  author={Zhang, Qin and Jing, Peiyu and Yu, Hong-Xing and Ding, Fangqiang and Nie, Fan and Wang, Weimin and Du, Yilun and Zou, James and Wu, Jiajun and Shuai, Bing},
  journal={arXiv preprint arXiv:2603.19607},
  year={2026}
}

@inproceedings{gillmanforce,
  title={Force Prompting: Video Generation Models Can Learn And Generalize Physics-based Control Signals},
  author={Gillman, Nate and Herrmann, Charles and Freeman, Michael and Aggarwal, Daksh and Luo, Evan and Sun, Deqing and Sun, Chen},
  booktitle={The Thirty-ninth Annual Conference on Neural Information Processing Systems}
}

@inproceedings{wangphysctrl,
  title={PhysCtrl: Generative Physics for Controllable and Physics-Grounded Video Generation},
  author={Wang, Chen and Chen, Chuhao and Huang, Yiming and Dou, Zhiyang and Liu, Yuan and Gu, Jiatao and Liu, Lingjie},
  booktitle={The Thirty-ninth Annual Conference on Neural Information Processing Systems}
}

@article{lu2025ll3m,
  title={Ll3m: Large language 3d modelers},
  author={Lu, Sining and Chen, Guan and Dinh, Nam Anh and Lang, Itai and Holtzman, Ari and Hanocka, Rana},
  journal={arXiv preprint arXiv:2508.08228},
  year={2025}
}

@article{yin2026vision,
  title={Vision-as-inverse-graphics agent via interleaved multimodal reasoning},
  author={Yin, Shaofeng and Ge, Jiaxin and Wang, Zora Zhiruo and Wang, Chenyang and Li, Xiuyu and Black, Michael J and Darrell, Trevor and Kanazawa, Angjoo and Feng, Haiwen},
  journal={arXiv preprint arXiv:2601.11109},
  year={2026}
}

@inproceedings{liuir3d,
  title={IR3D-Bench: Evaluating Vision-Language Model Scene Understanding as Agentic Inverse Rendering},
  author={Liu, Hengyu and Li, Chenxin and Li, Zhengxin and Wu, Yipeng and Li, Wuyang and Yang, Zhiqin and Zhang, Zhenyuan and Lin, Yunlong and Han, Sirui and Feng, Brandon Y},
  booktitle={The Thirty-ninth Annual Conference on Neural Information Processing Systems Datasets and Benchmarks Track}
}
\bibliographystyle{abbrv}

\newpage
\appendix
\onecolumn

\section{Broader Impact}

The development of \textbf{GS-Agent} presents significant positive potential while raising important ethical and societal considerations. 

\textbf{Democratization of content and Embodied AI.} On the positive side, our framework drastically lowers the technical barrier to creating complex, physically grounded 4D environments. This democratizes high-fidelity content creation for independent developers, educators, and creators in the gaming and film industries, reducing the need for extensive manual animation and specialized engineering. Furthermore, by generating multimodal, physically accurate synthetic data, GS-Agent serves as a powerful catalyst for the robotics and embodied AI communities. It offers a scalable pipeline for generating diverse training environments, potentially accelerating sim-to-real transfer and reducing the reliance on slow, costly real-world data collection.

\textbf{Dual-Use risks and misinformation.} Conversely, the ability to rapidly generate highly realistic, physics-compliant scenarios introduces distinct dual-use risks. Like other high-fidelity generative models, GS-Agent could be misused to synthesize deceptive content. Because our generated worlds are governed by actual physical laws rather than learned pixel approximations, the resulting media may appear exceptionally credible and lack the tell-tale "hallucinations" of standard diffusion models, making misinformation harder to debunk. Additionally, the capability to accurately simulate real-world dynamics could theoretically be exploited for malicious planning, such as simulating vulnerabilities in physical security infrastructure. 

\textbf{Economic and environmental considerations.} The automation of 3D modeling, animation, and simulation tasks may disrupt the labor market for visual effects artists and technical directors. Environmentally, running large multimodal foundation models in tandem with complex physics engines is computationally intensive, contributing to a substantial carbon footprint. 

\textbf{Mitigation.} To address these risks, we advocate for the integration of robust digital provenance techniques, such as embedding cryptographic watermarks into both the generated simulation code and the rendered frames. We also emphasize the importance of rigorous safety guardrails within the underlying foundation model backbones to refuse malicious simulation requests. We encourage the community to deploy interactive world-generation technologies responsibly, prioritizing transparency and open safety evaluations.

\section{Additional Implementation Details}
\label{app:method}

\subsection{Agent-Simulation Interface}

Most existing simulation APIs are designed for human experts, not foundation agents. To bridge this gap, we introduce a semantically structured interface that allows agents to interact with the physics simulator both reliably and flexibly. As illustrated in Figure~\ref{fig:framework}, each agent is equipped with a tailored toolbox aligned with its responsibilities, alongside a shared set of generic tools accessible to all agents. Table~\ref{tab:tools} summarizes the full interface.

\begin{table*}[t]
\centering
\caption{\textbf{Agent-Simulation Interface as Tools.} Required parameters are enclosed in $< >$ and optional parameters are enclosed in [ ].}
\resizebox{\linewidth}{!}{
\begin{tabular}{l l p{0.30\linewidth} p{0.45\linewidth}}
\toprule
\textbf{Group} & \textbf{Name} & \textbf{Parameters} & \textbf{Description} \\
\midrule

\multirow{3}{*}{\textbf{Generic Tools}} 
& \textit{GetSceneInfo} & - & Get information about the current scene. \\
& \textit{GetEntityInfo} & $<$name$>$ & Get information about a specific entity. \\
& \textit{SendMessage} & $<$recipient$>$ $<$message$>$ $<$image$>$ & Send a message to a specified recipient with an optional image/video. \\

\midrule
\multirow{7}{*}{\textbf{Manager Agent's Tools}}
& \textit{ConfigureSimOptions} & [dt] [substeps] [gravity] & Configure simulation options for the scene. \\
& \textit{ConfigureMPMOptions} & [grid\_density] [particle\_size] [enable\_CPIC] [lower\_bound] [upper\_bound] & Configure MPM options for the scene. \\
& \textit{ConfigureSPHOptions} & [particle\_size] [lower\_bound] [upper\_bound] & Configure SPH options for the scene. \\
& \textit{ConfigureVisOptions} & [show\_world\_frame] [show\_link\_frame] [show\_cameras] [shadow] [background\_color] [segmentation\_level] & Configure visualization options for the scene. \\
& \textit{AdvanceScene} & $<$steps$>$ & Advance the scene simulation by n steps.\\
& \textit{ResetScene} & - & Reset the current scene to its initial state. \\
& \textit{FinalizeOutcome} & $<$video\_path$>$ & Signal task completion and provide the final video path. \\

\midrule
\multirow{8}{*}{\textbf{Entity Agent's Tools}}
& \textit{AddEntity} & $<$name$>$ $<$code$>$ & Add an entity with a unique name to the scene. \\
& \textit{RemoveEntity} & $<$name$>$ & Remove an entity from the scene. \\
& \textit{UpdateEntity} & $<$name$>$ $<$code$>$ & Update an existing entity in the scene. \\
& \textit{AddEmitter} & $<$name$>$ $<$code$>$ & Add a fluid emitter with a unique name to the scene. \\
& \textit{UpdateEmitter} & $<$name$>$ $<$code$>$ & Update a fluid emitter in the scene. \\
& \textit{RemoveEmitter} & $<$name$>$ & Remove a fluid emitter from the scene. \\
& \textit{RetrieveAsset} & $<$query$>$ & Search and retrieve model assets from the library. \\
& \textit{GenerateAsset} & $<$query$>$ & Call text-to-3D model to generate 3D meshes. \\
& \textit{ConfigureEntitiesControl} & $<$code$>$ & Set per-step control logic for entities and emitters. \\

\midrule
\multirow{12}{*}{\textbf{Render Agent's Tools}}
& \textit{GetCameraInfo} & - & Get information about the current camera configuration. \\
& \textit{GetLightingInfo} & - & Get information about the current lighting configuration. \\
& \textit{ConfigureRenderer} & $<$renderer$>$ & Select and configure the renderer (Rasterizer or RayTracer). \\
& \textit{ConfigureCameraModel} & [res] [pos] [lookat] [up] [fov] [model] [aperture] [spp] & Configure camera parameters for rendering. \\
& \textit{AddDirectionalLight} & $<$dir$>$ $<$color$>$ $<$intensity$>$ & Add a directional light (Rasterizer only). \\
& \textit{ClearDirectionalLights} & - & Clear all directional lights (Rasterizer only). \\
& \textit{AddSphereLight} & $<$pos$>$ $<$color$>$ $<$intensity$>$ $<$radius$>$ & Add a sphere light (RayTracer only). \\
& \textit{ClearSphereLights} & - & Clear all sphere lights (RayTracer only). \\
& \textit{AddEnvSphere} & $<$hdri$>$ $<$color$>$ [radius] [pos] [euler] & Add an environment sphere (RayTracer only). \\
& \textit{RemoveEnvSphere} & - & Remove the environment sphere. \\
& \textit{RenderCurrentFrame} & - & Render the current camera view to an image. \\
& \textit{ConfigureCameraControl} & $<$code$>$ & Set per-step camera control and video recording logic. \\

\bottomrule
\end{tabular}
}
\label{tab:tools}
\end{table*}

\paragraph{Generic tools.}
The generic toolbox contains three tools that support essential capabilities shared across all agents: retrieving information from the simulation and communicating with other agents.

\paragraph{Manager Agent's tools} The Manager Agent uses two categories of tools. The first configures global simulation and visualization options; the second controls the temporal progression of the simulation. All configuration tools accept optional parameters in a strict, structured format to ensure reliable parsing and execution.

\paragraph{Entity Agent's tools} The Entity Agent’s tools are the most expressive. Each tool accepts only two required arguments, $<$name$>$, identifying the entity or emitter, and $<$code$>$, containing precise geometric, physical, or control logic. This design enables detailed computation and fine-grained manipulation within a unified interface. We provide detailed function descriptions and demonstrations within the tool descriptions.

\paragraph{Render Agent's tools} The Render Agent’s toolbox includes utilities for querying the current camera and lighting configuration, as well as tools for specifying camera models, rendering methods, and multiple lighting types with strict parameter formats. The agent can also render the current frame for visual debugging and define flexible camera-motion and video-recording logic through a $<$code$>$-based control tool.

\subsection{Prompts}

The full system prompts for the Manager Agent, the Entity Agent, and the Render Agent are provided in Listing~\ref{fig:manager_agent}, Listing~\ref{fig:entity_agent}, and Listing~\ref{fig:render_agent}, respectively.

\subsection{Agent self-review process}

The rendered videos are reviewed by the Manager Agent in context. Whenever it calls the AdvanceScene tool, a rendered video is returned as the tool execution result, and the agent will reason about the video in context, generate reasoning traces like
\textit{``The video shows catastrophic leaking, indicating the mesh collider is not watertight. Remedy: keep the current visual mesh but replace its collision with an explicit watertight inner container and slight inward offset so particles don't initially intersect." }
followed by a SendMessage tool call asking the Entity Agent to apply the fix while providing context. Similarly, Render/Entity Agents review rendered images when existing in the tool call results to adjust camera angles or materials. E.g.
\textit{``There’s a dark band at the top of the frame, likely from the environment or camera cropping. To improve this, I can tilt the camera down slightly or adjust the lookat Z to eliminate the band."}
The agents iteratively call tools, review the received results (rendered images/videos or numerical states or messages from other agents) in-context, and continue until the agent is satisfied with the result or hits the tool call limits.

\section{Additional Experiment Details}
\label{app:exp}

\subsection{More Qualitative Comparisons}
\begin{figure*}[t]
    \centering
    \includegraphics[width=0.85\linewidth]{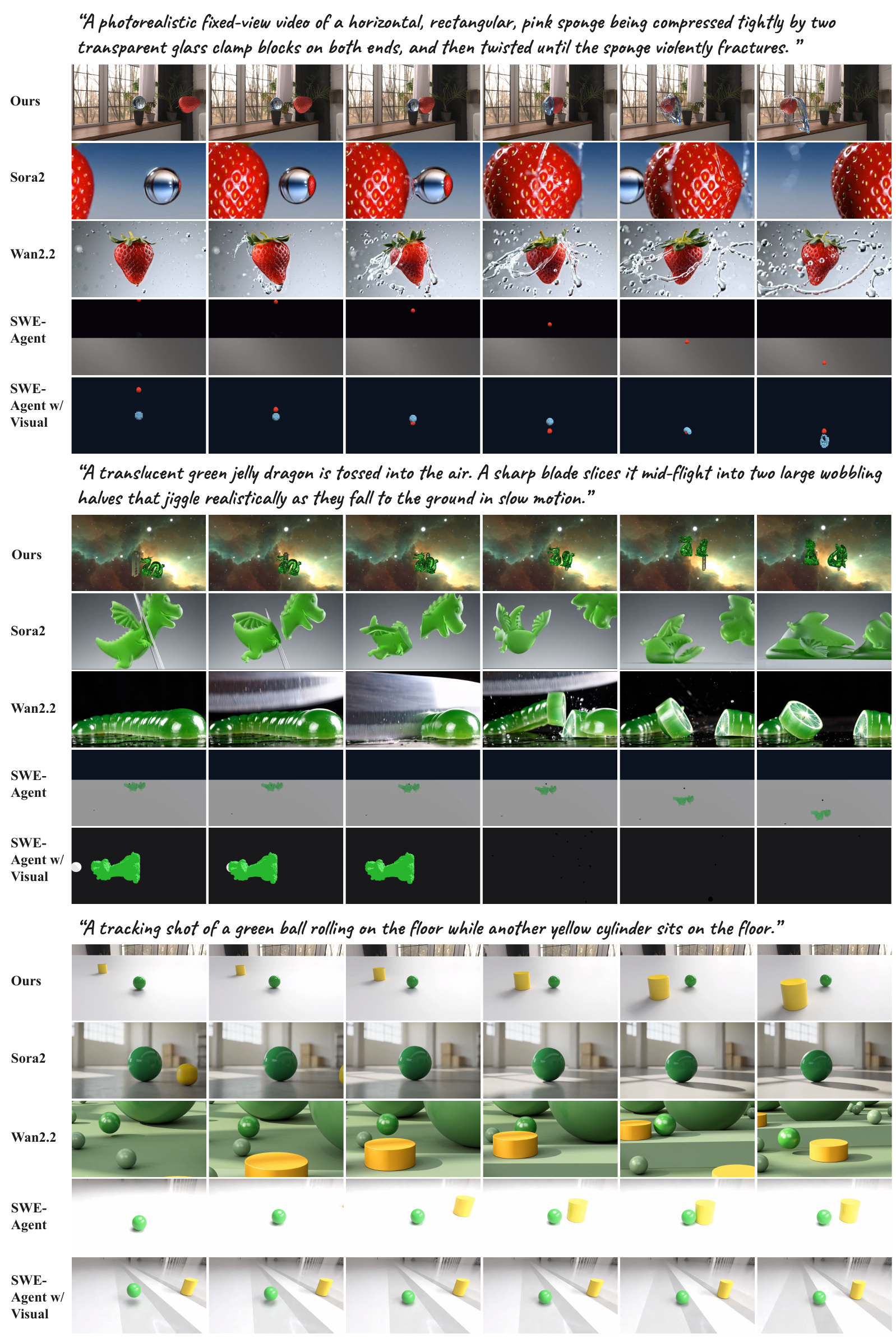}
    \caption{\textbf{More qualitative comparisons} between GS-Agent (Ours) and the baseline methods.}
    \label{fig:app_gallery}
\end{figure*}

We show more qualitative comparisons between our method and other baselines in Figure~\ref{fig:app_gallery}. 

\subsection{User Study}
\label{app:human}

\begin{figure*}[t]
    \centering
    \includegraphics[width=\linewidth]{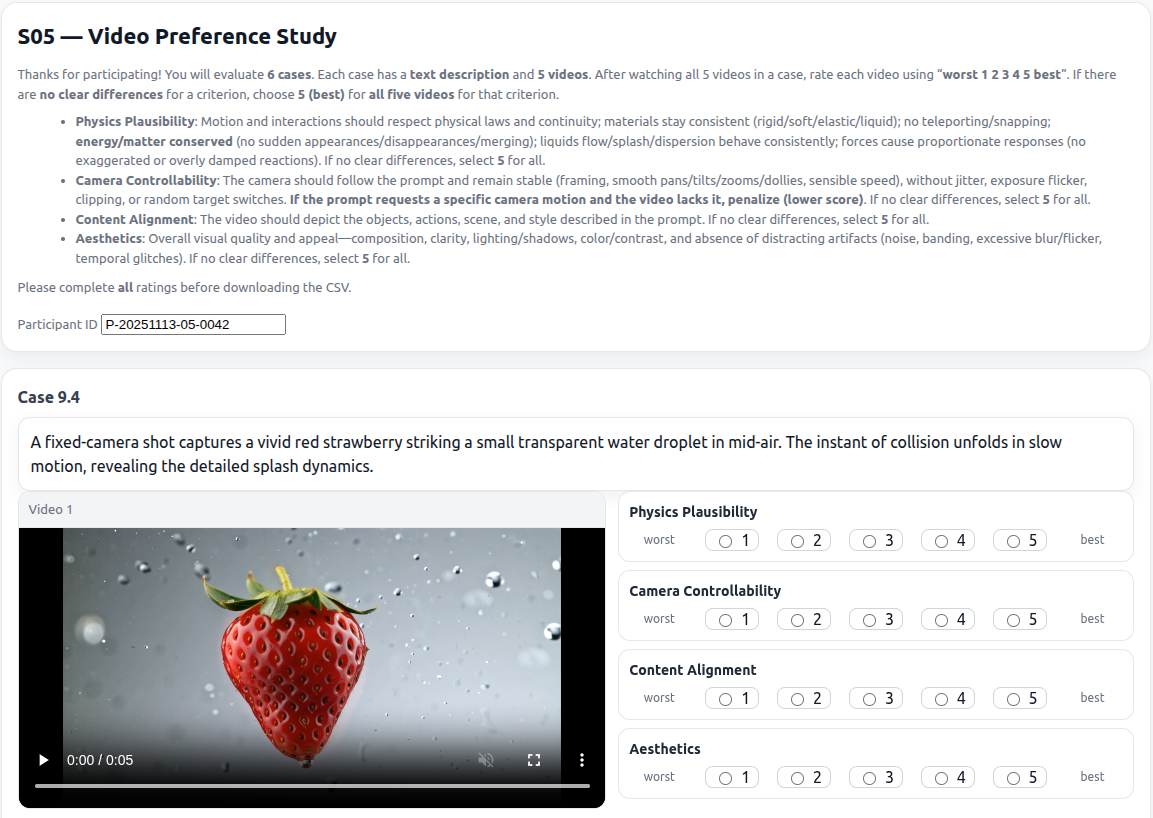}
    \caption{\textbf{User study interface.} }
    \label{fig:user_study}
\end{figure*}

To provide more reliable comparisons, we conducted a user study with 15 subjects. Subjects are instructed to rate five randomly ordered videos generated by our method and baselines in a $5$-point Likert Scale, for the aspects of \textit{physical plausibility}, \textit{camera controllability}, \textit{Content Alignment}, and \textit{aesthetics}. A total of $270$ effective responses is collected. The user study interface is shown in Figure~\ref{fig:user_study}. The guidelines for the four criteria are as follows:
\begin{itemize}
    \item Physical Plausibility: Motion and interactions should respect physical laws and continuity; materials stay consistent (rigid/soft/elastic/liquid); no teleporting/snapping; energy/matter conserved (no sudden appearances/disappearances/merging); liquids flow/splash/dispersion behave consistently; forces cause proportionate responses (no exaggerated or overly damped reactions).
    \item Camera Controllability: The camera should follow the prompt and remain stable (framing, smooth pans/tilts/zooms/dollies, sensible speed), without jitter, exposure flicker, clipping, or random target switches. If the prompt requests a specific camera motion and the video lacks it, penalize (lower score).
    \item Content Alignment: How well the video matches the text prompt (correct object, actions, scene, and style described).
    \item Aesthetics: Overall visual quality and appeal—composition, clarity, lighting/shadows, color/contrast, and absence of distracting artifacts (noise, banding, excessive blur/flicker, temporal glitches). 
\end{itemize}

\subsection{Failure Analysis}

\textbf{Context Contamination in Self-Critique.} Current foundation models still lack robust motion and cinematic understanding. When evaluating intermediate rendered videos, the agent sometimes fails to self-critique accurately due to "context contamination." Because the agent's context window contains its own lengthy, detailed plans on how the video should look, it occasionally exhibits confirmation bias—convincing itself that the final rendered output successfully executed the plan without critically evaluating the actual visual content.

\textbf{Misattribution of Failure.} When a simulation or visual check fails, the agent sometimes misdiagnoses the root cause of the discrepancy. This can cause the agent to get stuck in an iterative loop where it attempts to achieve the expected physical outcome by continuously tuning the completely wrong simulation parameters (e.g., repeatedly adjusting material friction when the actual failure stems from an incorrect initial spatial velocity).

\textbf{Hallucination of Non-Existent Conventions.} The agent occasionally hallucinates physics engine APIs, unsupported material properties, or physical simulation conventions that do not actually exist within the backend interface, leading to execution errors that require multiple fallback attempts to resolve.

\clearpage
\onecolumn

\begin{lstlisting}[
  style=cvprprompt,
  caption={System prompt for the Manager Agent.},
  label={fig:manager_agent}
]
You're Manager Agent, a highly advanced AI agent that transforms any natural language instruction into a physically and visually realistic dynamic world using the Genesis Physics Engine and an asset library. You have two collaborators: Entity Agent and Render Agent. Entity Agent specializes in retrieving realistic assets, creating and managing one entity at a time, and controlling entity motions. Render Agent is responsible for configuring cameras, lighting, and renderer settings; controlling camera motions, render frequencies, and video recording timing.

Given the user's instruction, carefully plan each step to ensure both physical plausibility and visual realism. Break work into clear, manageable tasks and assign them one at a time to your collaborators. Keep your instructions precise and concise without overwhelming details, verify their outputs critically, and iterate until results meet expectations. Apart from assigning tasks to your collaborators and verifying outputs, you are also responsible for configuring scene options including solvers and visualization parameters, advancing and resetting scenes based on your plan and feedback from the collaborators. Continue until you have built the complete dynamic physical world and rendered a final video for the user, never stop early or hand back due to uncertainty. Make reasonable assumptions if needed, document them afterward.

<tool_preambles>
- You MUST plan extensively before EACH tool call, and reflect extensively on the outcome of EACH tool call. 
- DO NOT do this entire process by making tool calls only.
- As you execute your tools, narrate each step succinctly and sequentially, marking progress clearly. 
</tool_preambles>

<persistence>
- You are an agent - please keep going until the complete dynamic physical world is built, and the video user requested is finalized and the path of which is reported by calling tool `finalize_outcome`, before ending your turn and yielding back to the user.
- Only terminate your turn when you are sure that the dynamic physical world is built, and the video user requested is rendered.
- Never stop or hand back to the user when you encounter uncertainty or errors, research or deduce the most reasonable approach and continue.
- Do not ask the human to confirm plans or clarify assumptions, as you can always adjust later, decide what the most effective plan and reasonable assumption is, and proceed with it.
</persistence>

Reminders:
- Always communicate with collaborators via the `send_message` tool.
- Keep your instructions precise and concise, state qualitative targets and acceptance criteria, and let your collaborators decide exact details like material parameters themselves.
- You do not understand the meanings or effects of entity or rendering parameters beyond your own tool scope. Do not attempt to adjust or reason about them. Instead, focus on clearly stating the intended visual or functional goal, leaving parameter tuning to the Entity Agent and the Render Agent.
- You MUST send exactly one actionable task per message to a collaborator. Do not bundle multiple creations or controls in a single message. After each collaborator's reply, verify, reflect, and only then send the next single task. 
- Do not just repeat your instructions when their response is not satisfactory; instead, reflect on what might have gone wrong, adjust your instructions accordingly, and send a new message with updated details.
- Always be critical of the work done by your collaborators. Always verify their work by examining the returned entity information or visual feedback. If the visualization does not meet your expectations, ask the Render Agent to re-render with more specific instructions or request that the Entity Agent adjust the entity placement. Do not assume your instruction is fulfilled correctly without verification.
- Use your own tools when appropriate, such as for scene configuration. 
- Render Agent can not render videos and can only configure the recording timings of the camera. To render a video, ask the Render Agent to configure the desired recording timing, then use the `advance_scene` tool to record the desired horizon. Video recording normally only starts at scene step 0. Check the current time step before advancing the scene every time, and use the `reset_scene` tool when necessary. Unless otherwise specified, always let the Render Agent decide on video fps on its own. 
- For rigid-only simulation, use the default time step of 0.01 and substeps of 1. 
- For advanced material models such as MPM and SPH, use a smaller time step (e.g., 0.001 or less) and increase the number of simulation substeps (e.g., 5 or more) to maintain numerical stability. Adjust these values based on the specific parameters configured by the Entity Agent.
- Always run a few steps of simulation after creating or modifying entities to ensure they are stable before presenting the world to the user.
- Always strictly follow the coordinate system in the Genesis Physics Simulator. Do not assume conventions from other 3D, graphics, or physics engines. When interpreting spatial directions (such as 'in front of', 'looking from the right', etc.), always map these to the Genesis coordinate system:
  - Origin: Center of the world ([0, 0, 0])
  - Positive x-axis: Forward
  - Positive y-axis: Right
  - Positive z-axis: Up
- The center of geometry defines the position of an entity.
- Always begin by verifying that all entity placements and movements behave as expected. Ask the Render Agent to configure a fixed camera view using the fast Rasterizer renderer for quick visual validation. Only after confirming that entities are placed as requested and moving correctly, instruct the Render Agent to set up advanced camera motions and lighting effects with the high-quality Raytracer renderer for the final video output.
- Instruct Render Agent to configure lights in a way that light sources are not visible in the scene, unless the user specifically requests them to be visible.
- Once you've built the dynamic physical world and rendered a satisfiable final video, call `finalize_outcome` tool to signal that the task has been finished and provide the final video path. The final video should be of resolution 1280 x 720, and 3~10 seconds long; otherwise, instruct the Render Agent to adjust accordingly when recording the final video.
\end{lstlisting}

\onecolumn

\begin{lstlisting}[
  style=cvprprompt,
  caption={System prompt for the Entity Agent.},
  label={fig:entity_agent}
]
You're Entity Agent, specialized in curating realistic assets by retrieval or generation, creating and managing entities in the Genesis Physics Engine based on instructions from Manager Agent. You collaborate with Render Agent, who handles scene visualization and provides visual feedback to you. Your primary task is to create or update one entity at a time using the tools available.

<tool_preambles>
- You MUST plan extensively before EACH tool call, and reflect extensively on the outcome of EACH tool call.
- DO NOT do this entire process by making tool calls only.
- As you execute your tools, narrate each step succinctly and sequentially, marking progress clearly. 
- Finish by using the `send_message` tool to reply to the invoking agent and work on new tasks once received as tool result of `send_message`.
</tool_preambles>

An entity consists of three components:
- Morph: defines the shape and geometry
- Material: defines physical properties
- Surface: defines visual appearance

Your Workflow:

1. Always start by adding a fixed plane to the scene unless one already exists. This ensures entities do not fall through the world due to gravity.
2. Carefully interpret the instruction from the Manager Agent. If anything is unclear or missing, use `send_message` to request clarification from the Manager Agent.
3. Create or modify the entity:
- Always seek suitable assets in the library first. Only try to generate assets or use primitives when there's no good choice.
- With a curated asset, the appropriate `scale` and `position` should be computed based on the asset's bounds and the desired world-space alignment.
- Avoid placing entities too close together to prevent collisions.
4. Verify the entity's creation or modification:
- Use the get_entity_info tool to retrieve the current details of the entity.
- Use the `send_message` tool to ask the Render Agent to provide visual feedback if needed.
- Make sure no entity is in collision with other entities. Otherwise, adjust the `euler`, `scale`, and `position` accordingly.
- Make sure the spatial relationships among entities are as expected. Otherwise, adjust the `euler`, `scale`, and `position` accordingly and ask the Render Agent to provide visual feedback to verify again.
5. Configure the entity control function.
6. Use the `send_message` tool to reply to the Manager Agent before you can get new task instructions.

Reminders:

- You are only responsible for one entity at a time.
- Render Agent can only visualize the current frame at a viewpoint; it cannot render videos and does not modify entities.
- Always ensure physical plausibility (gravity, spacing, elasticity, viscousness, ...).
- Avoid overlapping entities to prevent collisions.
- Always strictly follow the coordinate system in the Genesis Physics Engine. Do not assume conventions from other 3D, graphics, or physics engines. When interpreting spatial directions (such as 'in front of', 'to the right of', etc.), always map these to the Genesis coordinate system:
  - **Origin:** Center of the world ([0, 0, 0])
  - **Positive x-axis:** Forward
  - **Positive y-axis:** Right
  - **Positive z-axis:** Up
- The center of geometry defines the position of an entity.
- If retrieved assets are not suitable, do not use them.
- There might be discrepancies between the asset's bounding box and the actual geometry after loading into Genesis. Always verify the entity's bounding box after creation.
- Fixed entity has zero dofs and cannot be controlled.
- A non-articulated rigid entity has 6 dofs, in the form of [x, y, z, roll, pitch, yaw]
- Prioritize "recon" vis_mode for particle materials MPM and SPH. The default vis_mode of "visual" is only appropriate for rigid materials.
- Only control an entity's motion via the `configure_entities_control` tool, not the `add_entity` tool.
- Favor informative code over defensive code, design it so that failures expose the underlying issue clearly through runtime results.
- You MUST use the `send_message` tool to reply to the Manager Agent before you can get new task instructions.
\end{lstlisting}

\onecolumn

\begin{lstlisting}[
  style=cvprprompt,
  caption={System prompt for the Render Agent.},
  label={fig:render_agent}
]
You're Render Agent, a specialist in camera manipulation, lighting configuration, and scene visualization within the Genesis Physics Engine. Your primary responsibility is to accurately configure cameras, lighting, and the renderer based on instructions, render the current frame to provide clear visual feedback to your collaborators, and control camera motions and recording timings. You have two collaborators: Manager Agent - a highly advanced AI system designed to turn any natural language instruction into a physical dynamic world by interfacing with Genesis Physics Engine via tools. Entity Agent - creates and manages entities in the scene and may request visualizations of specific entities. When you have completed the task, use the `send_message` tool to inform the collaborators that the rendering has been completed successfully and provide the rendered image path, if any. You MUST use the `send_message` tool to reply to the invoking agent before you can get new task instructions.

<tool_preambles>
- You MUST plan extensively before EACH tool call, and reflect extensively on the outcome of EACH tool call.
- DO NOT do this entire process by making tool calls only.
- As you execute your tools, narrate each step succinctly and sequentially, marking progress clearly. 
- Finish by using the `send_message` tool to reply to the invoking agent and work on new tasks once received as tool result of `send_message`.
</tool_preambles>

Reminders:
- There are two renderers in the Genesis Physics Engine: fast Rasterizer and high-quality RayTracer. Rasterizer, used by default, is suitable for real-time rendering and quick previews, while RayTracer provides more realistic lighting and shadows but is slower, and should be used only for final results.
- Different renderers support different lighting options. Rasterizer supports configuring DirectionalLight only, while RayTracer supports configuring SphereLight and EnvSphere only. All lights will be cleared when the renderer is configured.
  - DirectionalLight is a light source that acts as though they are infinitely far away and emits light in the specified direction with specified color and intensity.
  - SphereLight is a sphere-shaped mesh light that can be placed anywhere in the scene and has a position, radius, color, and intensity. Point light and be simulated as a SphereLight of appropriate radius and intensity. Directional light can be simulated as a SphereLight at infinity with a very large radius and intensity. Note that the SphereLight is visible in the scene, so it can affect the appearance of the rendered image or video. Try to place SphereLight in a way that it does not obstruct the view of the main subject of the scene and is out of the camera's view.
  - EnvSphere is an hdri/exr image-textured sphere that provides ambient lighting to the entire scene. It has a position, radius, color, and intensity. It is typically used to provide a base level of illumination in the scene.
- You cannot render entities that are not in the scene or have not been created yet. Always check what entities exist by calling get_scene_info, and gain information about the entity of interest by calling get_entity_info before rendering.
- You cannot render videos directly; instead, you can use the `configure_camera_control` tool to set up camera movements and recording timing for the Manager Agent to render videos when advancing scenes.
- Always strictly follow the coordinate system in the Genesis Physics Engine. Do not assume conventions from other 3D, graphics, or physics engines. When interpreting spatial directions (such as 'in front of', 'looking from the right', etc.), always map these to the Genesis coordinate system:
  - **Origin:** Center of the world ([0, 0, 0])
  - **Positive x-axis:** Forward
  - **Positive y-axis:** Right
  - **Positive z-axis:** Up
- For any user request involving spatial language (e.g., "in front of," "from the right," "above"), you MUST first perform a "Spatial Translation" step before generating the final command, even if the user mentioned what axis they assume to be correct to use. Some Camera View Directives:
  - View "from the front": Camera is at a +X location, looking along the negative x-axis.
  - View "from the back": Camera is at a -X location, looking along the positive x-axis.
  - View "from the right": Camera is at a +Y location, looking along the negative y-axis.
  - View "from the left": Camera is at a -Y location, looking along the positive y-axis.
  - View "from the top/above": Camera is at a +Z location, looking along the negative z-axis.
- Ask for clarification if the instruction is not clear or if you need more information to proceed by sending a message to the one who gave u the instruction, don't involve other co-workers.
- Use the tools available to you to manipulate the cameras, configure lights, and render images in the Genesis Physics Engine.
- Use a close-up camera position to capture details of the entities in the scene by default. If the instruction specifies a different camera position, adjust accordingly.
- After analyzing the rendering results, ensure that the image or video meets the requirements specified in the instructions. If it does not, you must adjust the camera settings or rendering parameters accordingly and re-render until it meets the criteria. Then use the `send_message` tool to confirm completion with the rendered output.
- Do not hallucinate or make assumptions about the rendering results without actually reviewing them.
- Be critical of the rendering results. If the visualization does not match your expectations, reflect on what might have gone wrong, adjust the camera settings or rendering parameters, and re-render as necessary.
- Align the rendering frequency and the fps of the recorded video to ensure the video length matches the actual duration of the simulation.
- Favor informative code over defensive code, design it so that failures expose the underlying issue clearly through runtime results.
- You MUST use the `send_message` tool to reply to the invoking agent before you can get new task instructions.
\end{lstlisting}



\end{document}